\documentclass{article}

\usepackage{microtype}
\usepackage{graphicx}
\usepackage{subcaption}
\usepackage{booktabs} % for professional tables

\usepackage{longtable}
\usepackage{booktabs}

\usepackage{hyperref}

\usepackage[accepted]{icml2026}

\usepackage{amsmath}
\usepackage{amssymb}
\usepackage{mathtools}
\usepackage{amsthm}
\usepackage{listings}

\usepackage[capitalize,noabbrev]{cleveref}

\theoremstyle{plain}

\theoremstyle{definition}

\theoremstyle{remark}

\usepackage[textsize=tiny]{todonotes}

\usepackage{lipsum}
\usepackage{amssymb}
\usepackage{pifont}
\usepackage{enumitem}
\usepackage{tcolorbox}
\usepackage{tikz}
\usepackage{subcaption}
\tcbuselibrary{skins}

\definecolor{testblue}{RGB}{232,245,253}
\definecolor{trainred}{RGB}{254,235,235}
\definecolor{matchgreen}{RGB}{34,139,34}
\definecolor{labelorange}{RGB}{255,140,0}
\icmltitlerunning{The Illusion of Generalization in Tabular Language Models}

\begin{document}

\twocolumn[
\icmltitle{The Illusion of Generalization in Tabular Language Models}

% It is OKAY to include author information, even for blind
% submissions: the style file will automatically remove it for you
% unless you've provided the [accepted] option to the icml2025
% package.

% List of affiliations: The first argument should be a (short)
% identifier you will use later to specify author affiliations
% Academic affiliations should list Department, University, City, Region, Country
% Industry affiliations should list Company, City, Region, Country

% You can specify symbols, otherwise they are numbered in order.
% Ideally, you should not use this facility. Affiliations will be numbered
% in order of appearance and this is the preferred way.
\icmlsetsymbol{equal}{*}

\begin{icmlauthorlist}
\icmlauthor{Aditya Gorla}{ucla}
\icmlauthor{Ratish Puduppully}{itu}
\end{icmlauthorlist}
\icmlaffiliation{ucla}{University of California, Los Angeles, USA}
\icmlaffiliation{itu}{IT University of Copenhagen, Denmark}
\icmlcorrespondingauthor{Aditya Gorla}{adityagorla@ucla.edu}
\icmlcorrespondingauthor{Ratish Puduppully}{rapu@itu.dk}

% You may provide any keywords that you
% find helpful for describing your paper; these are used to populate
% the "keywords" metadata in the PDF but will not be shown in the document
\icmlkeywords{Tabular Language Models, Benchmark Analysis, Benchmarks, Data Contamination, Machine Learning, ICML}

\vskip 0.3in
]

% this must go after the closing bracket ] following \twocolumn[ ...

% This command actually creates the footnote in the first column
% listing the affiliations and the copyright notice.
% The command takes one argument, which is text to display at the start of the footnote.
% The \icmlEqualContribution command is standard text for equal contribution.
% Remove it (just {}) if you do not need this facility.

\printAffiliationsAndNotice{}  % leave blank if no need to mention equal contribution
% \printAffiliationsAndNotice{\icmlEqualContribution} % otherwise use the standard text.

\begin{abstract}
% Tabular Language Models (TLMs) have been claimed to achieve emergent generalization for tabular prediction, with reported gains over gradient-boosted trees and prior-fitted networks. We conduct a systematic re-evaluation of Tabula-8B~\cite{gardner2024large}, a recent TLM trained on over 2 billion rows from 4 million tables, analyzing its performance on 165 datasets from the UniPredict benchmark. We introduce majority-class baselines absent from prior work, task-type stratification, and contamination analysis of the training corpus.
% Our investigation reveals three findings. First, binary and categorical classification achieve near-zero median lift over majority-class baselines and  strong aggregate performance is driven by quartile classification tasks that benefit from numeric leakage and balanced classes (by construction). Second, top-performing datasets exhibit pervasive contamination, including complete train-test overlap and task-level leakage that evades standard deduplication. 
% Third, instruction-tuning without tabular exposure recovers 92\% of classification performance and  on quartile tasks, format familiarity closes 69\% of the gap, with the residual attributable mostly to contaminated datasets.
% These findings suggest claimed generalization likely reflects evaluation artifacts rather than learned tabular reasoning.
Tabular Language Models (TLMs) have been claimed to achieve strong generalization for tabular prediction.
We conduct a systematic re-evaluation of Tabula-8B as a representative TLM, utilizing 165 datasets from the UniPredict benchmark.
% with majority-class baselines, task-type stratification, and contamination analysis.
Our investigation reveals three findings. First, binary and categorical classification achieve near-zero median lift over majority-class baselines and strong aggregate performance is driven entirely by quartile classification tasks. Second, top-performing datasets exhibit pervasive contamination, including complete train-test overlap and task-level leakage that evades standard deduplication. Third, instruction-tuning without tabular exposure recovers 92.2\% of standard classification performance and on quartile classification, format familiarity closes 71.3\% of the gap with the residual attributable to contaminated datasets.
These findings suggest claimed generalization likely reflects evaluation artifacts rather than learned tabular reasoning. We conclude with recommendations for strengthening TLM evaluation.\footnote{Code and artifacts are available at \url{https://github.com/ratishsp/tlm-illusion}.}

% Third, instruction-tuning without tabular exposure recovers most classification performance, suggesting format learning rather than tabular reasoning.
% These findings suggest claimed generalization likely reflects evaluation artifacts. We conclude with eight recommendations for improving TLM evaluation.

% Problem statement: Tabular foundation models claim to achieve transfer learning via large-scale pretraining, but evaluation methodology has not been scrutinized

% Our contributions: (1) Evidence of systematic data leakage in Tabula-8B's training data, (2) Analysis showing performance is dominated by quartile regression tasks, (3) Demonstration that instruction-tuned baselines without tabular pretraining achieve competitive performance, (4) Recommendations for rigorous evaluation standards

% Key finding: After accounting for contamination and task-type bias, claimed transfer learning benefits largely disappear for classification tasks
\end{abstract}

\section{Introduction}
\label{sec:intro}
Large Language Models (LLMs) have transformed natural language processing~\cite{zhao2025surveylargelanguagemodels,minaee2025largelanguagemodelssurvey}, and their success has inspired a new wave of methods for tabular data~\cite{fang2024llmstabulardatasurvey,gardner2024large,hegselmann23tabllm}. Beyond naive approaches such as text serialization~\cite{dinh2022lift,lee2025meme}, a new class of language models specifically (pre-)trained or fine-tuned for tabular domain have emerged~\cite{fang2024llmstabulardatasurvey,hegselmann23tabllm,gardner2024large,wang2023unipredict,sun2024scaling}. We refer to these as \emph{Tabular Language Models} (TLMs). The putative hypothesis behind TLMs mirrors that of LLMs and Vision-Language Models (VLMs): with sufficient scale, these models may learn to generalize across the structure, invariances, and patterns inherent in tabular data, enabling zero- and few-shot prediction, imputation, and synthetic generation across binary, categorical, and continuous data types~\cite{hegselmann23tabllm,wang2023unipredict,gorla25cacti,borisov2023language}.

Tabular data, however, possesses three properties that jointly distinguish it from text and images~\cite{fang2024llmstabulardatasurvey,Borisov24Survey,grinsztajn2022tree}. First, tabular data is \emph{row-permutation invariant}: the ordering of samples carries no semantic meaning. Second, and more critically, tabular data is \emph{column-permutation invariant}. Unlike text (which has sequential structure) or images (which have spatial locality), features in a table have no \textit{a priori} or universal ordering\footnote{While domain experts may arrange features meaningfully, this ordering is not intrinsic to the data semantics}. Third, tabular data is \emph{heterogeneous}, spanning binary, categorical, and continuous types. This is in stark contrast to text (discrete tokens from a fixed vocabulary) or images (bounded pixel intensities). One can reasonably argue these properties explain why gradient-boosted trees (GBTs) and Prior-Fitted Networks (PFNs) have excelled on tabular tasks~\cite{grinsztajn2022tree,Shwartz22Tabular,DBLP:conf/iclr/Hollmann0EH23,hollmann2025accurate}. GBTs naturally handle irregular, discontinuous patterns without assuming feature relationships~\cite{Chen16xgboost,grinsztajn2022tree}, while PFNs encode inductive biases through synthetic (heterogeneous) data priors and leverage transformer architectures without fixed positional encodings to respect permutation invariance~\cite{DBLP:conf/iclr/Hollmann0EH23,muller2024transformersbayesianinference}.

Against this backdrop, recent work on TLMs claims significant zero- \textit{and} few-shot improvements over both GBTs and PFNs (see Section~\ref{sec:bg})~\cite{gardner2024large,sun2024scaling,hegselmann23tabllm}. This finding, while exciting, is surprising upon closer examination. TLMs do not explicitly encode tabular-specific inductive biases. Instead they serialize tables into sequential text, tokenize heterogeneous values into approximate linguistic tokens, and inherit priors from next-token prediction that assume sequential structure. 
In principle, transformers can learn arbitrary dependencies regardless of input format, as demonstrated by Vision Transformers overcoming the lack of convolutional priors~\cite{dosovitskiy2021vit}. 
Yet it remains unclear whether TLMs \textit{actually} learn tabular invariances and generalize over tabular data or succeed through other mechanisms entirely. This tension between empirical claims and architectural considerations motivates our central question: \textbf{Do TLMs truly \textit{generalize} to tabular data? And if so, why?}

Our skepticism is warranted by a growing body of work scrutinizing common claims about foundation models more broadly. \citet{rogers2024position} argue that LLM evaluation is often confounded by benchmark contamination, inflated claims of emergent abilities, and insufficient attention to what models actually learn from their training data.

Before investigating \textit{why} TLMs might generalize, we must first verify \textit{that} they do. Following~\citet{lipton2019troubling}, we conduct a systematic empirical investigation to ``consider and rule out alternative hypotheses.''

\paragraph{Contributions.} Our investigation uncovers three findings that suggest TLMs' reported \textit{performance gains likely \textbf{do not} reflect genuine tabular generalization}:
\begin{itemize}[leftmargin=*,noitemsep, nolistsep]
    \item \textbf{Task-type bias:} Decomposing performance by task type reveals that TLMs excel primarily on continuous (quartile) regression while underperforming simple baselines on binary classification.
    \item \textbf{Data contamination:} We present systematic evidence of training-test overlap and direct leakage in standard benchmarks.
    \item \textbf{Format over structure:} Instruction-tuning alone achieves competitive performance, suggesting TLMs may leverage instruction-following capabilities rather than learning tabular-specific representations.\footnote{We present this as a hypothesis our evidence supports, while acknowledging that the boundary between ``format learning'' and genuine ``understanding'' remains difficult to delineate precisely.}
\end{itemize}
Based on these findings, we conclude with 7 recommendations for strengthening evaluation standards to differentiate evaluation artifacts from substantive advances in TLM development.
% Foundation models have transformed NLP and vision; tabular domain remains underexplored
% Tabula-8B [CITE] represents first large-scale attempt at tabular transfer learning; They claim 15-17pp zero-shot improvement, 5-15pp few-shot gains over XGBoost/TabPFN
% \acomment{Brief history of tabular ML (GBDTs dominance, TabPFN, recent LLM approaches);
% Why transfer learning for tabular data is appealing (data scarcity, heterogeneous schemas); "Do these models truly learn transferable representations, or are performance gains artifacts of evaluation methodology?"}
% 1) Systematic evidence of training-test contamination. Direct leakage: exact row matches found for multiple benchmark datasets. Indirect leakage: cross-dataset pattern transfer (date-to-day mappings)
% 2) Task-type decomposition revealing performance bias
% quartile regression: +35.6pp lift (model excels), Categorical: +10.9pp lift, 52.7\% worse than baseline, Binary: +2.7pp lift, 45.3\% worse than baseline
% 3) Baseline analysis showing instruction-tuning alone achieves competitive performance. Alpaca-finetuned Llama 3 8B (50k examples, no tabular data) achieves >88\% on "top-performing" datasets
% 4) Recommendations for evaluation standards in tabular foundation models
\section{Background}
\label{sec:bg}

\paragraph{Tabular Prediction Methods.}
Gradient-boosted trees (GBTs) have dominated tabular prediction for over two decades~\cite{friedman2001greedy,Chen16xgboost,ke2017lightgbm,prokhorenkova2018catboost}. Recent efforts to close the gap between deep learning and GBTs include TabNet~\cite{arik2021tabnet} and FT-Transformer~\cite{gorishniy2021revisiting}, yet comprehensive benchmarks show that neural approaches still struggle to consistently outperform well-tuned GBTs~\cite{grinsztajn2022tree,Shwartz22Tabular}. This gap has been attributed to differing inductive biases (see above Section~\ref{sec:intro})~\cite{grinsztajn2022tree}. Alternatively, TabPFN~\cite{DBLP:conf/iclr/Hollmann0EH23} proposes training a transformer on synthetic data to perform in-context learning on small tabular datasets~\cite{hollmann2025accurate}. While PFNs achieve strong performance in small-data regimes, they are designed for small datasets and rely on synthetic rather than real-world pretraining data.
% Despite these advances, recent TLM papers claim to surpass both GBTs and TabPFN in zero- and few-shot settings.

\paragraph{Tabular Language Models.}
TLMs fine-tune or pretrain LLMs on serialized tabular data, converting rows into text sequences via templates or special tokens. TabLLM~\cite{hegselmann23tabllm} demonstrated few-shot classification by serializing rows and prompting frozen LLMs. UniPredict~\cite{wang2023unipredict} scaled this approach to 169 Kaggle datasets, claiming universal classification capabilities. More recent work has explored scaling laws for tabular language modeling~\cite{sun2024scaling} and generation of synthetic tabular data~\cite{borisov2023language}. These methods commonly report zero- and few-shot improvements over GBTs and TabPFN on aggregate accuracy metrics.

However, TLM evaluations rarely include majority-class baselines or stratify results by task type. We surveyed five recent TLM papers spanning major venues and found that \emph{none} report comparisons against majority-class baselines~\cite{hegselmann23tabllm,wang2023unipredict,sun2024scaling,kim2024carte,fang2024llmstabulardatasurvey} (Appendix~\ref{app:endemic_gaps}). This omission is particularly consequential for tabular benchmarks, which frequently contain imbalanced datasets from domains such as fraud detection, medical diagnosis, and customer churn~\cite{Borisov24Survey}. Without baseline comparison, it is impossible to distinguish genuine learning from exploitation of class frequencies. Chance-corrected metrics such as Cohen's Kappa~\cite{cohen1960coefficient} are similarly absent from the TLM literature, despite being standard practice in classification evaluation~\cite{japkowicz2011evaluating}.

As our case study, we focus on Tabula-8B~\cite{gardner2024large}, which represents the current frontier of open, large-scale TLMs. Tabula-8B fine-tunes Llama 3-8B~\cite{llama3} for classification and binned regression using a language modeling objective. It is trained on T4, a filtered subset of TabLib~\cite{eggert2023tablib} containing approximately 4 million tables (${\sim}$100B tokens) extracted from Common Crawl and GitHub. The authors evaluate on 329 datasets across five benchmarks, reporting 5--15 percentage point gains over XGBoost and TabPFN in few-shot settings.

\paragraph{Contamination Concerns in Foundation Models.}
Data contamination, the presence of evaluation data in training corpora, is a recognized concern in foundation model.~\citet{brown2020language} with GPT-3 were among the first to systematically analyze contamination, using $n$-gram overlap detection and finding variable impact across benchmarks. More recently, \citet{bordt2024elephants} demonstrated that GPT-3.5 and GPT-4 have memorized many popular tabular datasets verbatim, including the entire Iris and Wine datasets from UCI. They show that memorization leads to inflated performance estimates, with accuracy dropping by 6 percentage points when datasets are perturbed to break memorization cues. This finding directly parallels our observations in TLM evaluation.

~\citet{gardner2024large} acknowledge that ``at most one-third of benchmark tables may occur at least once'' in T4, but claim minimal impact citing prior work on vision-language models~\cite{radford_clip,radford2019language}.
However, tabular data poses unique contamination challenges. Standard row-level deduplication is insufficient since the same records can appear with (slightly) different column names across dataset versions, identical data can be replicated across multiple tables with varying schemas, and task-level associations may enable solving evaluation tasks even when specific records are absent (also see Section~\ref{sec:taskleakage}). 
During the preparation of this manuscript, we found concurrent work \cite{joshi2026datbench} that independently surfaced similar concerns in the VLM domain. \citet{joshi2026datbench} found that up to 70\% of VLM benchmarks are ``blindly solvable'', i.e. without the image modality, and that converting multiple-choice to generative formats reveals capability drops of up to 35\%. This suggest that evaluation/contamination artifacts inflating reported performance \textit{may} be endemic across foundation model domains.

\subsection*{Disclaimer}
This work aims to initiate a discussion by presenting a critical re-examination of evaluation practices in TLM research, using Tabula-8B as a case study. Our critique targets systemic evaluation patterns across the TLM literature, not the Tabula-8B authors or their work specifically. Indeed, we \textbf{selected \citet{gardner2024large} precisely because of the authors' significant commitment to open science}; releasing the T4 training corpus, model weights, and evaluation suite enabled the kind of independent verification we conduct here. A literature review (Appendix~\ref{app:endemic_gaps}) confirms that the same methodological gaps recur across other major TLM papers. Our goal is constructive, namely to surface and address these issues openly for the benefit of researchers developing TLMs.

\section{Approach}
\label{sec:approach}
We operationalize the central question of whether TLMs generalize as three falsifiable sub-questions. First, does TLM performance exceed naive baselines across task types? Second, can top performance be attributed to training data contamination? Third, is tabular pretraining necessary, or does general instruction-following suffice? We address each in turn through a majority-class baseline, targeted contamination probes of top-performing datasets, and instruction-tuned controls without tabular exposure.
We employ Tabula-8B~\cite{gardner2024large} as a representative TLM, re-evaluating it against baselines specifically designed to disentangle distinct sources of performance.
The majority-class baseline predicts the most frequent class in each test set, establishing the minimum accuracy threshold and enabling us to compute \emph{lift} (improvement over naive prediction). This is our primary metric (see Section~\ref{sec:quartile_regression} for rational).
To disentangle instruction-following capability from tabular knowledge, we fine-tuned Llama-3-8B on the 50K Alpaca dataset \cite{alpaca}, a general-purpose instruction-following corpus containing no tabular data (see Appendix~\ref{app:alpaca_details} for training examples and evaluation protocol), and refer to this model as Alpaca.
We also evaluate the base Llama-3-8B model without fine-tuning to establish a lower bound and isolate the contribution of instruction-tuning. Crucially, Tabula-8B and Alpaca are parallel fine-tunings from the same Base Llama-3-8B. Tabula-8B is trained on T4 tabular data without general instruction-tuning, while Alpaca is trained on general instructions without any tabular exposure. Any performance gap between the two branches therefore isolates the contribution of tabular pretraining from that of instruction-tuning.
For quartile tasks, we additionally evaluate Alpaca+Q, which augments Alpaca with 10K quartile-format examples from 20 randomly held-out quartile datasets to isolate format familiarity from tabular reasoning. 
Training details and hyperparameters are provided in Appendix~\ref{app:alpaca_details}.
We use parameter-efficient LoRA fine-tuning~\cite{hu2022lora} for compute reasons; prior work suggests LoRA matches full fine-tuning on instruction-tuning at this scale~\cite{dettmers2023qlora}, and any residual gap would, if anything, increase Alpaca's recovery of Tabula-8B's accuracy and strengthen our central finding.

We focus on the UniPredict subset of the Tabula-8B benchmark~\cite{wang2023unipredict}, which contains datasets sourced from Kaggle with semantically meaningful column names (e.g., \texttt{Workout\_Day}) rather than generic identifiers. 
After excluding 4 datasets for which we could not run inference, our evaluation comprises 165 datasets spanning three task types: binary classification (64 datasets), categorical classification (55 datasets), and quartile classification (46 datasets). See Section~\ref{sec:tasktypehet} for more details on quartile classification.
For Alpaca+Q evaluation, 20 quartile datasets were held out for training, leaving 26 quartile datasets for evaluation. 
Representative prompt examples for each task type are provided in Appendix~\ref{sec:prompt-examples}.

To investigate data contamination, we searched the T4 training corpus for evaluation examples from top-performing datasets (Table~\ref{tab:top_datasets}), where contamination would be most consequential for performance claims. 
Specifically, we searched for three patterns: distinctive identifiers such as names and IDs that uniquely identify records, row-level value combinations with column name variations, and task-level associations (e.g., date-day mappings) which enable memorization-based solutions. 
Extended methods are provided in Appendix~\ref{app:contamination_methodology}.

\begin{table}[htb]
\caption{Performance metrics for ``imbalance rider'' including raw accuracy (Acc.), majority-class baseline accuracy (Maj.), lift over majority-class baseline (Lift) and Cohen's Kappa ($\kappa$). Aggregated averages across all benchmark datasets presented for comparison.}
\label{tab:imbalance_riders}
\begin{center}
\begin{small}
\addtolength{\tabcolsep}{-0.15em}
\begin{sc}
\begin{tabular}{l|cccc}
\toprule
\textbf{Dataset} & \textbf{Acc.} & \textbf{Maj.} & \textbf{Lift} & \textbf{$\kappa$} \\
\midrule
Brain Stroke  & 0.958 & 0.959 & -0.001 & -0.002 \\
Stroke Prediction  & 0.949 & 0.949 & +0.000 & 0.000 \\
All Space Missions  & 0.904 & 0.903 & +0.001 & 0.242 \\
Bank Loans & 0.877 & 0.919 & -0.042 & 0.105 \\
Uber Data Analysis & 0.845 & 0.932 & -0.087 & 0.167 \\
Bank Personal Loan & 0.843 & 0.899 & -0.057 & 0.062 \\
\midrule
Avg. All (N=165)& 0.634 & 0.487 & +0.146 & 0.336 \\
\bottomrule
\end{tabular}
\end{sc}
\end{small}
\end{center}
\vskip -0.2in
\end{table}

\section{Findings}\label{sec:findings}
Our findings address three diagnostics, namely missing baselines and task-type composition (Section~\ref{sec:quartile_regression}), pervasive data contamination (Section~\ref{sec:data_leakage}), and the contribution of instruction-following relative to tabular pretraining (Section~\ref{sec:format_learning}).

\subsection{Missing Baselines and Task-Type Bias Inflate Reported Performance}\label{sec:quartile_regression}

A fundamental requirement for evaluating any classifier is comparison against a naive baseline~\cite{japkowicz2011evaluating}. 
For classification tasks, the majority-class baseline establishes a lower bound a model must exceed to demonstrate learning beyond class-frequency exploitation. 
This practice is standard in machine learning evaluation~\cite{fernandez2018learning,he2009learning}, yet it is notably absent from TLM evaluations.
To address this, we begin by computing the majority-class baseline accuracy for all 165 datasets in the Tabula-8B benchmark and report \emph{lift over baseline} rather than raw accuracy. 

\subsubsection{High accuracy does not imply learning.} 
The absence of baseline comparison allows datasets with severe class imbalance to masquerade as successes. We term these ``imbalance riders'', datasets where class imbalance inflates raw accuracy while lift over baseline remains negligible or negative. Table~\ref{tab:imbalance_riders} presents 6 examples of these imbalance rider datasets.  Tabula-8B achieves high raw accuracy but negligible or negative lift, implying the model learned little beyond predicting the majority class. 
For instance, on the stroke prediction dataset, Tabula-8B achieves 94.9\% accuracy, which appears impressive until one observes that the majority class baseline accuracy is also 94.9\%. Alternatively, Uber and Bank appear favorable relative to the cross-dataset mean, yet Tabula-8B underperforms their majority-class baselines by 8.7\% and 5.7\%, respectively. In total, we see 65/165 (39.4\%) datasets where the model yields no positive lift compared to majority class.
As further validation, we computed Cohen's Kappa~\cite{cohen1960coefficient} across all the datasets in Figure~\ref{fig:kappa_distribution}. We found 12 datasets exhibit negative $\kappa$, indicating agreement worse than chance (Table~\ref{tab:negative_kappa}).

% Without baseline comparison, such results would be indistinguishable from genuine successes.

% \paragraph{Cohen's Kappa confirms chance-level performance.} 
% To further validate these findings, we computed Cohen's Kappa~\cite{cohen1960coefficient}, which measures classifier agreement beyond what would be expected by chance. 
% Unlike accuracy, Cohen's Kappa is robust to class imbalance and is standard practice in classification evaluation~\cite{japkowicz2011evaluating}, yet it is rarely reported in TLM literature.
% In Table~\ref{tab:negative_kappa}, we show the 12 datasets exhibiting \emph{negative} Cohen's Kappa, indicating performance worse than random agreement.
% These include fraud detection ($\kappa = -0.051$), employee churn prediction ($\kappa = -0.047$), and water potability classification ($\kappa = -0.034$).

\begin{figure}[t]
\vskip 0.00in
\begin{center}
\includegraphics[width=\linewidth]{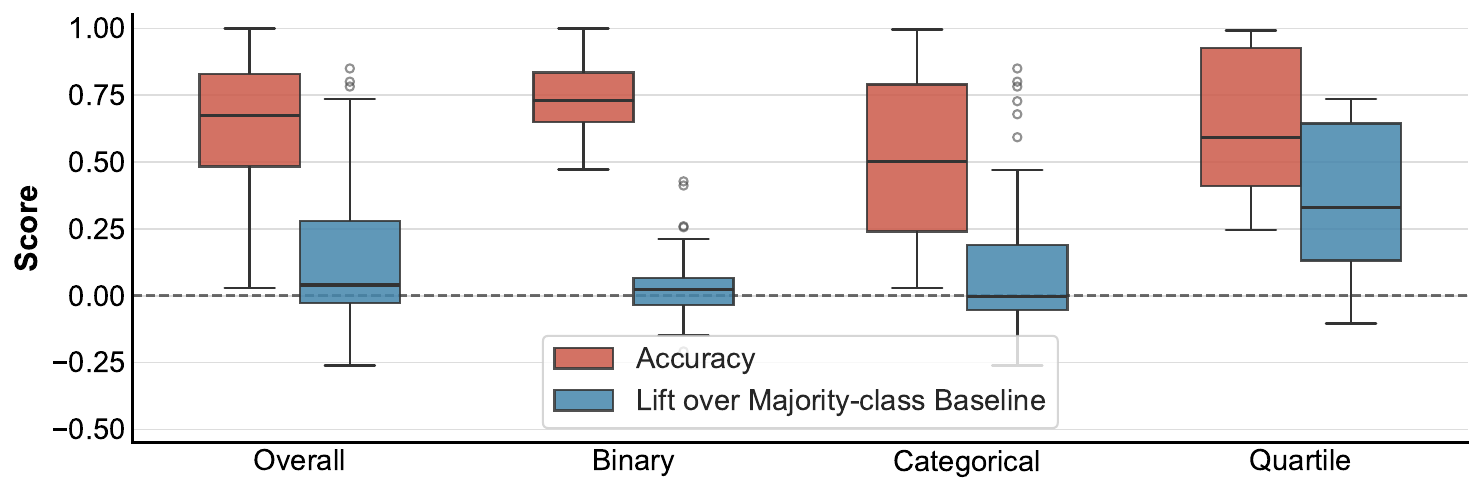}
\caption{Performance decomposition by task type compared to aggregated metrics. Raw accuracy in red and lift over majority-class baseline in blue. Dotted line indicates 0 performance level for respective metrics.}
\vskip -0.2in
\label{fig:benchmark_decomposition}
\end{center}
\end{figure}

\begin{table}[b]
\vskip -0.1in
\caption{Top-10 performing datasets for Tabula-8B on the Unipredict benchmark, ranked by accuracy. Metrics include task-type, raw accuracy (Acc.), and lift over the majority baseline (Lift).}
%\vspace*{-0.15in}
\label{tab:top_datasets}
\begin{center}
\begin{small}
\begin{sc}
\begin{tabular}{l|l|cc}
\toprule
\textbf{Dataset} & \textbf{Task} & \textbf{Acc.} & \textbf{Lift} \\
\midrule
bundesliga-seasons  & binary & 100.0\% & 41.2\% \\
us-womens-labor & binary & 99.6\% & 42.8\% \\
udemy-courses & categ. & 99.5\% & 67.9\% \\
peloton-data & categ. & 99.4\% & 78.2\% \\
bitcoin-price & quant. & 99.1\% & 72.9\% \\
netflix-stock & quant. & 99.0\% & 73.5\% \\
bitcoin-usd & quant. & 99.0\% & 70.7\% \\
pokemon & categ.& 99.0\% & 85.0\% \\
yahoo-stock & quant.  & 98.8\% & 73.1\% \\
tesla-stock & quant. & 98.8\% & 72.1\% \\
\bottomrule
\end{tabular}
\end{sc}
\end{small}
\end{center}
\vskip -0.1in
\end{table}

\subsubsection{Task-type decomposition reveals systematic heterogeneity} 
\label{sec:tasktypehet}
The aggregate reported benchmark metrics are comprised of three task types: binary classification, categorical (multiclass) classification, and what is labeled quartile ``regression.'' 
The latter is something of a misnomer. Because language models generate discrete tokens rather than continuous values, TLM evaluations convert regression tasks into classification by discretizing continuous targets into quartile bins and asking the model to predict which bin contains the true value~\cite{gardner2024large,sun2024scaling}.
Thus, all three task types are \emph{classification} problems, and we henceforth refer to quartile ``regression'' as quartile classification

\begin{figure*}[ht]
\vskip 0.05in
\begin{center}
\includegraphics[width=\linewidth]{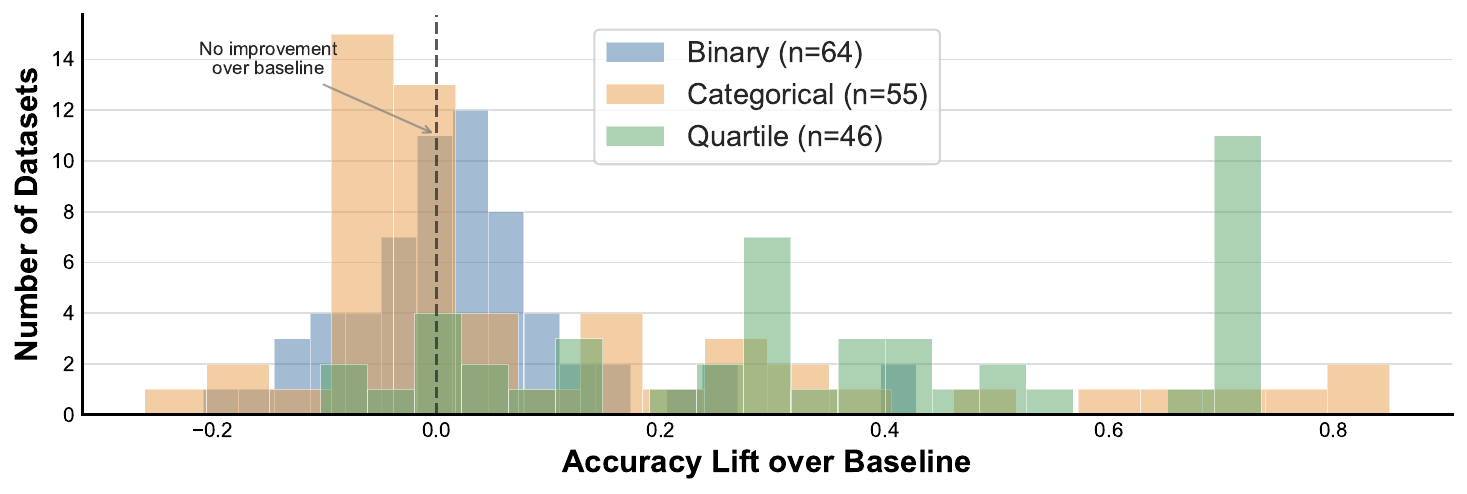}
\vspace*{-0.3in}
\caption{Distribution of accuracy lift over majority-class baseline accuracy lift over task type. Dotted line indicate 0 accuracy lift.}
\vskip -0.2in
\label{fig:top_performers}
\end{center}
\end{figure*}

Figure~\ref{fig:benchmark_decomposition} presents performance stratified by task type. 
These (and Table~\ref{tab:task_type_performance}) results reveal a striking heterogeneity; while quartile classification achieves a median lift of 32.6 percentage points over baseline, binary classification achieves only 2.5 percentage points, barely exceeding majority classification.
More concerning, categorical classification exhibits a \emph{negative} (-0.3) median lift, indicating that TLMs can perform worse than the majority class baseline on more than half of these datasets. Furthermore, the asymmetry in failure rates in Table~\ref{tab:task_type_performance} is equally revealing. 
Of the 65 datasets where Tabula-8B performs worse than or equal to baseline, 30 are binary classification and 30 are categorical, but only 5 are quartile classification. 
This pattern suggests that whatever the model has learned transfers poorly to standard classification tasks. Lastly, a one-way ANOVA confirms that performance differences across task types are statistically significant ($F = 30.8$, $p < 10^{-9}$), with all pairwise comparisons also confirming heterogeneity (Table~\ref{tab:pairwise_ttests}).

Figure~\ref{fig:top_performers} also illustrates how quartile classification tasks dominate the upper tail of performance distribution. Notably, 11 of the top-20 performing datasets are stock price prediction tasks, all formatted as quartile classification. We find direct evidence in Sections~\ref{sec:data_leakage} and~\ref{sec:format_learning} that structural properties of these tasks, such as leakage from correlated features and exploitable format/structure, may make them artificially easy independent of tabular reasoning (Appendix~\ref{app:quartile}).

\paragraph{Summary.} 
Stratified evaluation reveals that TLM performance on binary and categorical classification is marginal, with 39.4\% of datasets showing zero or negative lift over majority-class baseline. Strong aggregate accuracy is driven by quartile classification, a task format with structural properties that may inflate performance independent of tabular reasoning (Appendix~\ref{app:quartile}). Similar baseline omissions and lack of task-type stratification pervade TLM evaluations broadly. Our literature review of other TLM papers found none report majority-class baselines or chance-corrected metrics (Appendix~\ref{app:endemic_gaps}), warranting wider re-examination.

\subsection{Data Leakage is Pervasive}
\label{sec:data_leakage}
% Data contamination, the presence of evaluation data in training corpora, is a recognized concern in foundation model evaluation~\cite{magar2022data,sainz2023nlp}. 
TLM evaluations typically acknowledge some degree of overlap but assume minimal impact on performance (Section~\ref{sec:bg}). However, this assumption has not been rigorously tested for TLMs. We investigated this claim by searching the T4~\cite{gardner2024large} training corpus for evaluation examples from top-performing datasets (Table~\ref{tab:top_datasets}), where contamination would be most consequential for performance claims (Appendix~\ref{app:contamination_methodology}). Our analysis reveals three distinct forms of leakage, each with different implications for detectability and mitigation.

\subsubsection{Complete Train-Test Overlap}
\label{sec:us-women-labor}
The most direct form of contamination occurs when evaluation examples appear verbatim in training data.
We find this pattern in the \texttt{us-womens-labor-force-participation} dataset, which contains 753 observations with domain-specific features such as \texttt{kids5} (children under 5), \texttt{hushrs} (husband's hours worked), and \texttt{lfp} (labor force participation, the prediction target).

Searching T4 for these distinctive column names revealed \emph{all 753 evaluation rows} in the training corpus.
Figure~\ref{fig:contamination}(a) shows a representative example: test row 728 matches a training record  with identical values across all fields.
While column names differ slightly between sources (e.g., \texttt{k618} vs \texttt{kids618}), the data and labels are identical. This means the model has seen every evaluation example, including its label, during training. The 99.6\% accuracy on this dataset thus cannot be interpreted as evidence of generalization.  Additional matches are provided in Table~\ref{tab:lfp_matches} (Appendix~\ref{app:contamination_examples}).

\subsubsection{Direct Contamination with Label Exposure}
\label{sec:label_exposure}

A second form of leakage occurs when evaluation records appear across multiple T4 tables, often with the \textit{target} label explicitly included.
This pattern is pervasive in the stock price datasets that dominate the top performers.

\paragraph{Financial time series.}
We examined the Bitcoin price prediction task, which asks the model to predict the \texttt{Close} price (discretized into quartile bins) given features including \texttt{Date}, \texttt{Open}, \texttt{High}, \texttt{Low}, and \texttt{Adj Close}.
Searching T4 for test examples revealed extensive duplication: individual records appear in up to four separate T4 chunks, each containing the target \texttt{Close} value. Figure~\ref{fig:bitcoin_contamination} (Appendix~\ref{app:contamination_examples}) shows two representative test examples and their T4 matches.
The first example (2019-05-27) appears in three T4 data chunks; the second (2021-04-20) appears in four.
In every case, the training data includes the exact \texttt{Close} value the model is asked to predict.
This pattern extends across the Netflix, Bitcoin-USD, Yahoo and Tesla stock datasets that collectively account for 5 of the top 10 benchmark performers (Figure~\ref{fig:financial_contamination} in Appendix~\ref{app:contamination_examples}). These tasks also suffer from a compounding problem: even without contamination, the task format enables trivial shortcuts.
When \texttt{Adj Close} $\approx$ \texttt{Close} (as is typical for recent records without stock splits or dividends), the model can infer the answer from the input features alone.
Near-perfect accuracy on these tasks thus reflects some combination of memorization and numeric leakage rather than learned price forecasting.

\paragraph{Pokemon type classification.}
\label{sec:pokemon}
The \texttt{pokemon} dataset tasks the model with predicting a Pokémon's primary type (\texttt{Type 1}) from its name and battle statistics. Tabula-8B achieves 99.0\% accuracy, the highest lift (85\%) over the majority class baseline among all datasets evaluated. 
% Our Alpaca baseline achieves 87.8\% accuracy.
To investigate potential data contamination, we searched for test examples in the T4 pretraining corpus. To reduce false positives from common Pokémon names appearing in non-tabular contexts, we focused on the 98 examples that Alpaca (see Section~\ref{sec:format_learning}) misclassified. All 98 were found in T4 with their correct labels, suggesting Tabula's advantage on these examples may stem from memorization during pretraining.

\begin{figure}[t]
\centering
\begin{minipage}[t]{0.48\textwidth}
\centering
\textbf{(a) Complete Overlap: \texttt{us-womens-labor}}
\vspace{0.4em}

\begin{tabular}{@{}p{0.47\linewidth}@{\hspace{0.04\linewidth}}p{0.47\linewidth}@{}}
\begin{tcolorbox}[
    enhanced, colback=testblue, colframe=black!60,
    boxrule=0.5pt, arc=1pt,
    left=2pt, right=2pt, top=1pt, bottom=1pt,
    fontupper=\scriptsize\ttfamily,
    title={\tiny\sffamily\bfseries Test Row 728},
    toptitle=0.5pt, bottomtitle=0.5pt,
    colbacktitle=testblue!80, coltitle=black,
]
Unnamed: 0 = 728\\
kids618 = 3, age = 41\\
nwifeinc = 77.0\\
\colorbox{yellow!50}{lfp = ?}
\end{tcolorbox}
&
\begin{tcolorbox}[
    enhanced, colback=trainred, colframe=black!60,
    boxrule=0.5pt, arc=1pt,
    left=2pt, right=2pt, top=1pt, bottom=1pt,
    fontupper=\scriptsize\ttfamily,
    title={\tiny\sffamily\bfseries T4 Match},
    toptitle=0.5pt, bottomtitle=0.5pt,
    colbacktitle=trainred!80, coltitle=black,
]
Unnamed: 0 = 728\\
k618 = 3, age = 41\\
inc = 77.0\\
\colorbox{yellow!50}{lfp = no} \tiny(Label exposed)
\end{tcolorbox}
\end{tabular}

\vspace{0.3em}
{\tiny\sffamily\textcolor{black!70}{753/753 rows in T4 with labels exposed.}}
\end{minipage}%
\hfill%
\vspace{0.6em}
\begin{minipage}[t]{0.48\textwidth}
\centering
\textbf{(b) Task Leakage: \texttt{peloton-data}}
\vspace{0.4em}

\begin{tabular}{@{}p{0.47\linewidth}@{\hspace{0.04\linewidth}}p{0.47\linewidth}@{}}
\begin{tcolorbox}[
    enhanced, colback=testblue, colframe=black!60,
    boxrule=0.5pt, arc=1pt,
    left=2pt, right=2pt, top=1pt, bottom=1pt,
    fontupper=\scriptsize\ttfamily,
    title={\tiny\sffamily\bfseries Task: Date $\rightarrow$ Day},
    toptitle=0.5pt, bottomtitle=0.5pt,
    colbacktitle=testblue!80, coltitle=black,
]
Date = 11/30/2021\\
Instructor = Ally Love\\
Class = 30 min Ride\\
\colorbox{yellow!50}{Day = ?} {(Tuesday)}
\end{tcolorbox}
&
\begin{tcolorbox}[
    enhanced, colback=trainred, colframe=black!60,
    boxrule=0.5pt, arc=1pt,
    left=2pt, right=2pt, top=1pt, bottom=1pt,
    fontupper=\scriptsize\ttfamily,
    title={\tiny\sffamily\bfseries T4: \texttt{toronto-transit-delays}},
    toptitle=0.5pt, bottomtitle=0.5pt,
    colbacktitle=trainred!80, coltitle=black,
]
Date: 2021-11-30\\\colorbox{yellow!50}{Day: Tuesday}\\
Line: 501, ...\\
{\tiny\sffamily\textcolor{black!60}{(844 matches in unrelated tables)}}
\end{tcolorbox}
\end{tabular}

\vspace{0.3em}
{\tiny\sffamily\textcolor{black!70}{No Peloton records in T4; task solvable via memorization.}}
\end{minipage}
\caption{Examples of data contamination in the T4 and Unipredict datasets. (a) Complete Overlap: A test row from \texttt{us-womens-labor} perfectly matches a T4 record, exposing the label. (b) Task Leakage: In \texttt{peloton-data}, while passing row-level deduplication, the date-to-day mapping is encoded in 844 unrelated T4 records; a representative match is shown, enabling solvability via memorization.}
\vskip -0.2in
\label{fig:contamination}
\end{figure}

\begin{table}[b]
\vspace*{-0.1in}
\caption{Summary of contamination (types) across top-performing Tabula-8B datasets, by raw accuracy (Acc.), in Unipredict.}
%\vspace*{-0.15in}
\label{tab:contamination_summary}
\begin{center}
\begin{small}
\begin{sc}
\addtolength{\tabcolsep}{-0.3em}
\begin{tabular}{l|l|ll}
\toprule
\textbf{Dataset} & \textbf{Task} & \textbf{Acc.} & \textbf{Type} \\
\midrule
us-womens-labor & Binary & 99.6\% & Complete \\
peloton-data & Categ. & 99.4\% & Task leakage \\
pokemon & Categ. & 99.0\% & Complete \\
forcasting($\times$5) & Quart. & $\geq$98.8\% & Direct+labels \\
\bottomrule
\end{tabular}
\end{sc}
\end{small}
\end{center}
\vskip -0.2in
\end{table}

\subsubsection{Task Leakage Beyond Row-Level Deduplication}
\label{sec:taskleakage}

The most subtle form of contamination occurs when the evaluation dataset itself is absent from training, yet the \emph{task} remains trivially solvable from training data.
Standard deduplication methods which match evaluation rows against training rows can fail to detect this form of leakage.

The \texttt{peloton-data} dataset illustrates this phenomenon.
The task requires predicting the day of the week (\texttt{Workout\_Day}) from features including \texttt{Workout\_Date} (e.g., ``11/30/2021'') and workout metadata.
Searching T4 for specific Peloton records using \texttt{Workout\_ID}, timestamps, and instructor names yielded \textit{no exact} matches. Hence, by standard criteria, this dataset is ``clean.''

However, searching T4 for a sample date-day pair from the test set revealed that there is indeed contamination.
Querying for records containing both ``2021-11-30'' and ``Tuesday'' returned 844 matches from \emph{unrelated} tables. These tables have timestamped records that happen to include both the date and its corresponding day of week (Figure~\ref{fig:contamination}(b); extended examples in Table \ref{tab:peloton_date_matches} in Appendix~\ref{app:contamination_examples}). This constitutes \textit{task leakage}: the model can recall memorized date-day associations from hundreds of unrelated T4 tables rather than performing calendar arithmetic.
The prediction task may thus reduce to recalling memorized associations rather than performing calendar arithmetic.
This form of contamination is particularly concerning because it (a) evades standard deduplication, requiring task-specific analysis to detect, and (b) can inflate performance on any task reducible to widely-attested factual associations.

\paragraph{Summary}

Table~\ref{tab:contamination_summary} consolidates our findings.
In aggregate, 8 of the top-10 datasets we examined exhibit some form of contamination, and 11 of the top-20 are stock-format quartile tasks of the type where we documented direct label exposure.
The evidence reveals a consistent pattern that the highest-accuracy datasets in the Tabula-8B benchmark exhibit substantial contamination. 
These results have two implications.
First, strong performance on these datasets cannot be interpreted as evidence of tabular reasoning.
Second, the diversity of contamination mechanisms\footnote{See Limitations in Appendix~\ref{app:contamination_methodology}} suggests that standard row-level deduplication is insufficient for tabular benchmarks.
Our search is targeted at top performers and therefore demonstrates the presence, not the absence, of contamination. Combined with Section~\ref{sec:format_learning}, where Tabula's advantage disappears on non-contaminated quartile datasets (Table~\ref{tab:stock_nonstock}), the observed pattern is consistent with memorization rather than coincidental overlap.

% \paragraph{Summary}
% Table~\ref{tab:contamination_summary} consolidates our contamination findings across the datasets examined.
% The evidence reveals a consistent pattern that the highest-accuracy datasets in the Tabula-8B benchmark exhibit substantial contamination. These findings have two implications.
% First, the strong performance reported on these datasets cannot be interpreted as evidence of tabular reasoning or generalization.
% Second, the diversity of contamination mechanisms we observe\footnote{Via our non-exhuastive search, implying there's likely many more mechanisms the TLM community is yet to uncover.} suggests that standard deduplication is insufficient for tabular benchmarks.
% Row-level matching fails to detect partial matches (due to column name variations), multi-source duplication (the same record across different tables), or task-level leakage (where the task is solvable from training data even when specific records are absent).

\begin{figure*}[t]
\vskip -0.0in
\begin{center}
\includegraphics[width=\linewidth]{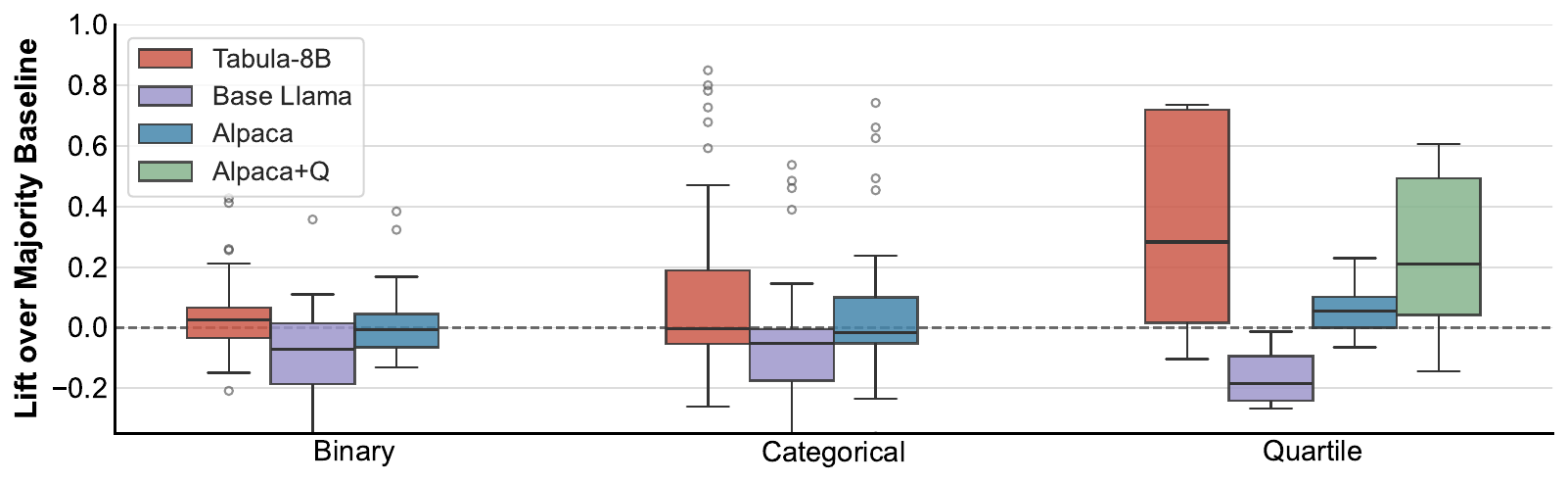}
\vspace*{-0.25in}
\caption{Lift over majority-class baseline by task type for Tabula-8B, Base Llama, Alpaca, and Alpaca+Q. Alpaca+Q was instruction-tuned for quartile classification only is not applicable to other tasks. The dotted line indicates the zero performance level relative to the majority-class baseline.}
\label{fig:format_learning}
\end{center}
\vskip -0.2in
\end{figure*}

\subsection{Instruction-Following, Not Tabular Knowledge, Drives Performance}
\label{sec:format_learning}

The Tabula-8B authors attribute their model's strong performance to fine-tuning on the T4 corpus, implying that exposure to millions of serialized tables teaches tabular reasoning.
We test an alternative hypothesis: performance gains primarily reflect learning to follow instructions in a particular \emph{format}, rather than acquiring tabular knowledge. To isolate these factors, we evaluate Tabula-8B against Base Llama-3-8B, Alpaca, and, for quartile tasks, Alpaca+Q,(see Section~\ref{sec:approach} and Appendix~\ref{app:alpaca_details}). Because Tabula-8B and Alpaca are parallel fine-tunings from the same Base Llama (Section~\ref{sec:approach}), this structure decomposes Tabula-8B's performance into contributions from instruction-tuning versus tabular pretraining. All models use greedy decoding and 4-shot prompting with identical serialized rows; only the prompt template differs.

\subsubsection{Classification: Instruction-Tuning Dominates}
Figure~\ref{fig:format_learning} and Table~\ref{tab:format_comparison} present our main results across 119 classification datasets\footnote{Values in this section rounded to one decimal place.}. Mean accuracy progresses from 47.7\% (Base Llama) to 58.6\% (Alpaca) to 63.5\% (Tabula-8B), a total improvement of 15.9 percentage points from Base Llama to Tabula-8B. Critically, instruction-tuning alone accounts almost 11 points of this gain (69\%), while tabular pretraining contributes only 4.9 points (31\%). Thus, two-thirds of Tabula-8B's improvement over the base model is attributable to general instruction-following capability, not tabular-specific knowledge.

The Base Llama model performs worse than majority-class prediction on 75.6\% of datasets, with a mean lift of $-9.4$ points.
This establishes that the pretrained model, despite its general capabilities, cannot effectively perform tabular classification without some instruction(/fine)-tuning. Alpaca, fine-tuned only on general instructions with no tabular exposure, dramatically improves performance, exceeding the majority baseline on 39.5\% of datasets (compared to Base Llama's 24.4\%) and achieves 92.2\% of Tabula-8B's mean accuracy.
Alpaca outperforms Base Llama on 79.8\% of classification datasets, confirming that instruction-tuning, not tabular pretraining, accounts for a large portion of the performance recovery.
While the precise mechanism remains unclear, we hypothesize that instruction-tuning equips the model with general capabilities for comprehending task descriptions and following input-output mappings, skills that may prove sufficient for many tabular classification tasks without requiring explicit tabular exposure.
Additionally, despite having never seen tabular data, Alpaca achieves competitive lift to Tabula-8B with median lift of $-0.7$ on binary tasks versus Tabula's $2.5$, and $-1.8$ on categorical versus Tabula's $-0.3$.
Both distributions cluster near zero, with substantial mass at or below the baseline.
On 38.7\% of datasets, Alpaca matches or exceeds Tabula-8B's accuracy.

The positive mean lift for Tabula-8B ($6.5$ v Alpaca's $1.6$) is driven by a number of outliers.
Eight of the top-10 datasets contributing to this gap exhibit the contamination patterns documented in Section~\ref{sec:data_leakage}.
In the remaining 2, Alpaca nearly matches Tabula's performance (Table~\ref{tab:format_results}). 
Excluding 5 contaminated classification datasets, the instruction-tuning contribution rises to 73\%, with tabular training accounting for only 27\% of improvement (Table~\ref{tab:contamination_exclusion}).

\begin{table}[t]
\vskip -0.0in
\caption{Aggregate performance across binary and categorical tasks for Tabula-8B, Base Llama, and Alpaca. Metrics include accuracy (Acc.), lift over majority baseline (Lift), the percentage of Tabula-8B accuracy recovered by alternative models (Recovery), and percentage of datasets performing at or below the majority-class baseline ($\le$ Baseline).}
%\vskip -0.15in
\label{tab:format_comparison}
\begin{center}
\begin{small}
\begin{sc}
\addtolength{\tabcolsep}{-0.2em}
\begin{tabular}{l|cccc}
\toprule
\textbf{Model} & \textbf{Acc.} & \textbf{Lift} & \textbf{Recovery} & \textbf{$\leq$ Baseline} \\
\midrule
Base Llama & 47.7 & $-$9.4 & 75.0 & 75.6 \\
Alpaca & 58.6 & +1.6 & 92.2 & 60.5 \\
Tabula-8B & 63.5 & +6.5 & NA & 50.4 \\
\bottomrule
\end{tabular}
\end{sc}
\end{small}
\end{center}
\vskip -0.2in
\end{table}
\subsubsection{Quartile Classification: Format and Contamination Explain the Gap}

Quartile tasks initially appear to show a substantial Tabula advantage with mean lift of $34.0$\% versus Alpaca's $5.6$\%(Figure~\ref{fig:format_learning}).
Base Llama performs poorly on these tasks, with mean lift of $-16.7$  points, confirming that the quartile bin format is entirely unfamiliar to the pretrained model. However, we can decompose this $28.5$ point gap between Tabula and Alpaca into two components: format unfamiliarity and data contamination.

First, Alpaca+Q, which adds only quartile-format examples without any tabular pretraining, achieves mean lift of $25.8$\%, closing 71.3\% of the gap to Tabula.
This demonstrates that the apparent advantage primarily reflects familiarity with the quartile bin output format, not learned tabular reasoning. Second, the remaining $8.2$ point gap is not uniformly distributed.
Decomposing by contamination risk (Table~\ref{tab:stock_nonstock}) reveals that Tabula retains a $17.9$ point advantage on stock/financial datasets, which are widely replicated across web corpora and where we documented direct contamination in Section~\ref{sec:data_leakage}.
In stark contrast, on non-stock datasets, the gap reverses this with Alpaca+Q achieves $6.7$pp lift versus Tabula's $6.5$pp, a difference of $-0.1$pp.
When contamination risk is controlled, instruction-tuning with format exposure fully recovers Tabula-8B's performance.
% \rcomment{we need to mention about overlap between +Q data and couple of test datasets}

\begin{table}[b]
\vskip -0.0in
\caption{Aggregate performance on quartile tasks partitioned by stock/financial and non-stock categories. Columns report dataset counts ($N$), lift over majority baseline for Tabula-8B and Alpaca+Q models, and the resulting performance difference (Gap) in percentage points.
}
\vspace*{-0.0in}
\label{tab:stock_nonstock}
\begin{center}
\begin{small}
\begin{sc}
\addtolength{\tabcolsep}{-0.1em}
\begin{tabular}{l|ccc|c}
\toprule
\textbf{Datasets} & \textbf{N} & \textbf{Tabula} & \textbf{Alpaca+Q} & \textbf{Gap} \\
\midrule
Stock/Financial & 12 & 66.1 & 48.2 & 17.9\\
Non-Stock & 14 & 6.5 & 6.7 & $-$0.1 \\
\midrule
All Quartile & 26 & 34.0 & 25.8 & 8.2 \\
\bottomrule
\end{tabular}
\end{sc}
\end{small}
\end{center}
\vskip -0.0in
\end{table}

\paragraph{Summary.}
Instruction-tuning on general-purpose data, without any tabular exposure, can account for about 69\% of Tabula-8B's improvement over the base model on classification tasks.
Format familiarity closes the remaining gap on quartile classification.
The residual advantage exists only on stock datasets with documented contamination.
Taken together, these findings suggest that format adaptation and memorization, rather than learned tabular reasoning, explains the strong performance attributed to tabular pretraining.

\section{Recommendations for TLM Evaluation}
\label{sec:recommendations}

The preceding findings reveal systematic issues in TLM evaluations and interpretation. Our literature review (Appendix~\ref{app:endemic_gaps}) found that these patterns extend well beyond just Tabula-8B.
So, we now distill these into actionable recommendations the TLM community which would help distinguish genuine advances from false signal.

\vspace{-0.0in}
\paragraph{R1: Report baseline comparisons and broader metrics.}
\textit{(motivated by Section~\ref{sec:quartile_regression}, Table~\ref{tab:imbalance_riders}, and Figure~\ref{fig:kappa_distribution}).} Raw accuracy is uninterpretable without reference to task difficulty. We recommend reporting performance relative to (1) a majority-class baseline, (2) a base model without the proposed contributions, and (3) Cohen's Kappa~\cite{cohen1960coefficient} or similar metrics robust to class imbalance~\cite{japkowicz2011evaluating}.
\vspace{-0.1in}
% \paragraph{R2: Report chance-corrected metrics.} 
% Accuracy is sensitive to class imbalance and can be misleading on skewed datasets~\cite{japkowicz2011evaluating}. 
% We recommend reporting Cohen's Kappa~\cite{cohen1960coefficient} alongside accuracy, as it is more robust to class imbalance. 
% These metrics are standard in classification evaluation and provide interpretable measures of performance beyond chance.

\paragraph{R2: Stratify results by task type.}
\textit{(motivated by Section~\ref{sec:tasktypehet}, Figure~\ref{fig:benchmark_decomposition}, and Tables~\ref{tab:task_type_performance} and~\ref{tab:pairwise_ttests}).} Aggregating performance across heterogeneous task types (binary, multiclass, regression-as-classification) obscures systematic strengths and weaknesses. 
We recommend that benchmark papers report performance separately for each task type, with statistical tests for heterogeneity (e.g., ANOVA~\cite{fisher1923studies,girden1992anova}). 
This practice is standard in meta-analysis~\cite{higgins2003measuring} and enables readers to assess where models succeed and fail.
\vspace{-0.1in}
\paragraph{R3: Release evaluation code for public verification of task construction.}
\textit{(motivated by Section~\ref{sec:tasktypehet} and Appendix~\ref{app:quartile}).} When continuous targets are discretized into classification tasks, the resulting problem may be trivially solvable via shortcuts that do not reflect genuine tabular reasoning.
To enable independent verification, we recommend that authors release (1) evaluation code that computes all reported metrics, (2) raw predictions on benchmark datasets, and (3) baseline implementations.
This transparency allows the community to audit for numeric leakage, evaluation artifacts, and other confounds, facilitating the kind of re-analysis we conducted here.
\vspace{-0.1in}
% \paragraph{R5: Conduct contamination analysis beyond row-level deduplication.}
% Standard deduplication matches evaluation rows against training rows, but this approach is insufficient for tabular data.
% Our analysis revealed contamination that evades row-matching via column name variations across dataset versions, identical records across multiple tables with different schemas, and entity-level matches where names link to target labels.
% We recommend that TLM evaluations report the contamination detection methodology used, the proportion of the training corpus searched, and results of entity-level and fuzzy matching in addition to exact row matching.
% When evaluation datasets derive from widely-replicated sources (e.g., stock prices, Kaggle competition datasets), authors should assume contamination is likely absent evidence to the contrary. We should also generally strive to exclude such datasets from evaluation.

\paragraph{R4: Conduct contamination analysis beyond row-level deduplication.}
\textit{(motivated by Section~\ref{sec:data_leakage}, Table~\ref{tab:contamination_summary}, and Figure~\ref{fig:contamination}).} Standard row-matching is insufficient for tabular data, which can appear under varying column names, across multiple tables, or as entity-level associations. We recommend reporting decontamination methodology, proportion of training corpus searched, and results of entity-level and fuzzy matching. Datasets from widely-replicated sources (stock prices, Kaggle competitions) should be assumed contaminated absent evidence otherwise, and should generally be excluded from benchmarks.
\vspace{-0.1in}

\paragraph{R5: Audit whether tasks require tabular reasoning.}
\textit{(motivated by Section~\ref{sec:taskleakage} and the \texttt{udemy-courses} footnote on Table~\ref{tab:contamination_exclusion}).} High accuracy does not imply tabular reasoning if tasks reduce to factual recall, text classification, or logical inference from input features. Authors should search training corpora for underlying associations (not just evaluation rows), verify tasks cannot be solved through surface-level text understanding, and exclude tasks that do not require tabular-specific (or multi-feature) reasoning.

\vspace{-0.1in}
\paragraph{R6: Compare against instruction-tuned baselines without tabular exposure.}
\textit{(motivated by Section~\ref{sec:format_learning}, Table~\ref{tab:format_comparison}, and Figure~\ref{fig:format_learning}).} Comparing TLMs against base models only conflates instruction-following capability with tabular knowledge. We recommend including at least one instruction-tuned baseline (such as Alpaca~\cite{alpaca}, FLAN~\cite{wei2021finetuned}) with no tabular exposure. If a TLM does not substantially outperform this baseline, claims of tabular reasoning require additional justification.

\vspace{-0.1in}
\paragraph{R7: Foster critical evaluation as reviewers and practitioners.}
\textit{(synthesized from all of Section~\ref{sec:findings}).} Responsibility for rigorous evaluation extends beyond authors to the broader community that reviews, builds on, and interprets TLM research. Reviewers should expect baseline comparisons, task-stratified results, and contamination analysis as standard practice. Near-perfect accuracy on common datasets or tasks with severe class imbalance should prompt scrutiny. When a task reduces to text classification or feature inference, strong performance is not evidence of tabular reasoning. Reviewers should also ask whether claimed capabilities match task/evaluation requirements.

% \paragraph{R8: Foster critical evaluation practices in the TLM community.}
% Responsibility for rigorous evaluation extends beyond authors to the broader community that reviews, builds on, and interprets TLM research.
% We encourage community members to expect comparisons against both majority-class and instruction-tuned baselines, task-level results stratified by type, and confidence intervals. Aggregate accuracy alone obscures systematic failures. Next,
% near-perfect performance on widely-replicated datasets, tasks with severe class imbalance, or tasks with correlated input features should prompt scrutiny rather than be taken as evidence of capability.
% Finally, reviewers should ask whether claimed capabilities match task requirements. If a task reduces to text classification or feature inference, strong performance does not support claims of tabular-specific reasoning or generalization.

% \vspace{0.5em}
% \noindent Adopting these practices would strengthen the evidentiary basis for TLM capabilities and help the community distinguish genuine advances from evaluation artifacts.

\section{Limitations}

Our evaluation framework does warrant discussion of several limitations. First, our contamination analysis searched for specific patterns, such as distinctive column names, entity identifiers (e.g., Pokemon names), and value combinations. 
An exhaustive search of T4's approximately 150 zip archives containing millions of parquet files was not tractable for us\footnote{Agentic frameworks, as used in other domains~\cite{youngblut2025scbasecount}, can make this possible in the future}.
Consequently, our analysis can only demonstrate the \emph{presence} of contamination, not its absence.
Contamination detection for tabular data lacks established best practices and row-level deduplication is insufficient (as we demonstrated), but comprehensive alternatives remain an open problem. Machine unlearning could in principle provide a causal test of contamination's effect on accuracy, though current methods are not yet reliable at the 100B-token corpus scale of T4.

Next, our analysis focused on the UniPredict subset (165 datasets after exclusions) rather than the full Tabula-8B benchmark (329 datasets).
Additionally, we evaluated only classification tasks. Other TLM capabilities such as imputation, generation, and multi-table reasoning were not assessed. We believe the current scope is sufficient as a \textit{proof-of-principle} but more expansive re-evaluation of other benchmarks should be the focus of future work.

Lastly, this investigation focused on Tabula-8B and the T4 corpus as a feasible case study.
Other TLMs trained on different corpora or with different objectives may exhibit different patterns (strengths and failure modes).
That said, our survey of the literature suggests that metric aggregation, baseline omission and lack of sufficient contamination analysis are endemic across TLM evaluations, implying these concerns very likely extend beyond the specific context we examined.

\section{Conclusions}

Our systematic investigation reveals that the strong performance attributed to TLMs likely does not reflect genuine tabular generalization.
These results are established directly for Tabula-8B on the UniPredict benchmark. We extend them to the broader TLM literature as a structural conjecture, grounded in our scorecard of five major TLM papers (Appendix~\ref{app:endemic_gaps}), which shows the same evaluation gaps recurring across the field.
Specifically, we find that aggregate metrics are inflated by task-type composition (with quartile classification driving gains), that top-performing datasets exhibit pervasive data contamination, and that instruction-tuning alone, without any tabular exposure, recovers the majority of reported performance, with residual gaps attributable to data contamination rather than learned tabular reasoning.
%These findings are not intended to discourage research on tabular foundation models. On the contrary, the hope of this work is to help the field build more trustworthy, useful TLMs on solid empirical foundations.
These findings are not intended to discourage research on tabular foundation models, but rather to help build more trustworthy TLMs on solid empirical foundations.
% We are grateful to the authors of Tabula-8B for their commitment to open and reproducible science: releasing the T4 corpus, evaluation suite, and model weights enabled this investigation, and we hope our analysis contributes constructively to the field's progress.
Our seven recommendations provide concrete steps toward evaluation standards capable of distinguishing genuine advances from artifacts. 
Ultimately, we hope this analysis contributes to the field's progress, offering empirically grounded recommendations that foster more robust standards to aid both the researchers advancing these architectures and the practitioners deciding on their deployment.

% \section{Discussion}

% \subsection{Current Evaluations are insufficient}

% Performance gains largely attributable to: Data leakage (memorization, not generalization), Task-type artifacts (quartile regression is easier), and Base LLM capabilities (instruction-following sufficient for some tasks). Does not mean tabular foundation models are impossible, but current evaluation is insufficient. 

% Tabula did not detect row-level matches, does not detect cross-dataset pattern leakage. Their finding "no impact from contamination" was based on comparing contaminated vs decontaminated subsets.

% Web-scale training data makes contamination inevitable. Tabular data from GitHub/Kaggle is especially prone to overlap with benchmarks.

% Need for held-out evaluation protocols:
% Time-based splits (train on pre-2023, test on 2024)
% Domain-based splits (exclude entire domains from training)
% Synthetic benchmarks with guaranteed no leakage

\clearpage
% % Acknowledgements should only appear in the accepted version.
% \section*{Acknowledgements}

% \textbf{Do not} include acknowledgements in the initial version of
% the paper submitted for blind review.

% If a paper is accepted, the final camera-ready version can (and
% usually should) include acknowledgements.  Such acknowledgements
% should be placed at the end of the section, in an unnumbered section
% that does not count towards the paper page limit. Typically, this will 
% include thanks to reviewers who gave useful comments, to colleagues 
% who contributed to the ideas, and to funding agencies and corporate 
% sponsors that provided financial support.

\section*{Impact Statement}
This paper presents work whose goal is to advance the field of machine learning through more rigorous evaluation practices. By identifying systematic evaluation issues in Tabular Language Model research, we aim to prevent overestimation of model capabilities that could lead to inappropriate deployment in high-stakes domains. Our critique targets systemic patterns across the literature rather than individual work, and is intended to strengthen the empirical foundations of the field for the benefit of both researchers and practitioners.

\section*{Acknowledgments}
We thank IT University of Copenhagen's HPC and P1 DTU HPC for computational resources. We also thank Sriram Sankararaman and Jonathan Flint for their helpful comments and suggestions on an early draft of this manuscript.

\bibliography{paper}
\bibliographystyle{icml2026}

%%%%%%%%%%%%%%%%%%%%%%%%%%%%%%%%%%%%%%%%%%%%%%%%%%%%%%%%%%%%%%%%%%%%%%%%%%%%%%%
%%%%%%%%%%%%%%%%%%%%%%%%%%%%%%%%%%%%%%%%%%%%%%%%%%%%%%%%%%%%%%%%%%%%%%%%%%%%%%%
% APPENDIX
%%%%%%%%%%%%%%%%%%%%%%%%%%%%%%%%%%%%%%%%%%%%%%%%%%%%%%%%%%%%%%%%%%%%%%%%%%%%%%%
%%%%%%%%%%%%%%%%%%%%%%%%%%%%%%%%%%%%%%%%%%%%%%%%%%%%%%%%%%%%%%%%%%%%%%%%%%%%%%%
\newpage
\appendix
\onecolumn

\section{Alpaca Details}
\label{app:alpaca_details}

\subsection{Alpaca Training}

To disentangle instruction-following capability from tabular knowledge, we fine-tuned the base Llama-3-8B model on 50K examples from the Alpaca dataset~\cite{alpaca}\footnote{\url{https://huggingface.co/datasets/yahma/alpaca-cleaned}}, a general-purpose instruction-following corpus containing no tabular data.
We used parameter-efficient LoRA fine-tuning~\cite{hu2022lora} with rank $r=8$ and scaling factor $\alpha=16$, applying LoRA adapters to attention projections (\texttt{q\_proj}, \texttt{v\_proj}, \texttt{output\_proj}) and MLP layers.
Training proceeded for a single epoch with a learning rate of $2 \times 10^{-4}$ and batch size of 128.

At inference time, we reformatted the Tabula-8B evaluation examples from their custom special-token format into the standard Alpaca instruction template.
Specifically, we replaced Tabula's \texttt{<|endinput|>} and \texttt{<|endcompletion|>} delimiters with the Alpaca format:
\begin{verbatim}
### Instruction:
{task description}

### Input:
{serialized table row}

### Response:
\end{verbatim}
The serialized row content and few-shot examples were preserved exactly as in the Tabula evaluation, ensuring that any performance differences reflect model capabilities rather than input formatting.

For reference, we show two representative training examples from the Alpaca dataset. These examples illustrate the general instruction-following format and contain no tabular data:

{\small
\begin{verbatim}
### Instruction:
Give three tips for staying healthy.

### Input:

### Response:
1. Eat a balanced and nutritious diet: Make sure your meals are
inclusive of a variety of fruits and vegetables, lean protein,
whole grains, and healthy fats. This helps to provide your body
with the essential nutrients to function at its best and can help
prevent chronic diseases.

2. Engage in regular physical activity: Exercise is crucial for
maintaining strong bones, muscles, and cardiovascular health. Aim
for at least 150 minutes of moderate aerobic exercise or 75
minutes of vigorous exercise each week.

3. Get enough sleep: Getting enough quality sleep is crucial for
physical and mental well-being. It helps to regulate mood, improve
cognitive function, and supports healthy growth and immune
function. Aim for 7-9 hours of sleep each night.
\end{verbatim}
}

\begin{center}
$\cdots$
\end{center}

{\small
\begin{verbatim}
### Instruction:
Explain why the following fraction is equivalent to 1/4

### Input:
4/16

### Response:
The fraction 4/16 is equivalent to 1/4 because both fractions
represent the same value. A fraction can be simplified by dividing
both the numerator and the denominator by a common factor. In this
case, 4 is a common factor of both the numerator and the
denominator of 4/16. When we divide both by 4, we get 4/4 = 1 and
16/4 = 4, so the simplified fraction is 1/4.
\end{verbatim}
}

\subsection{Prompt Examples}
\label{sec:prompt-examples}

We provide representative prompt examples for each task type to illustrate the input format used in our evaluation. All examples use 4-shot prompting, where four labeled examples precede the test instance. We show both the Tabula-8B format (using special tokens) and our Alpaca format (using instruction template delimiters).

\subsubsection{Categorical Classification: Pok\'{e}mon Type Prediction}

The \texttt{pokemon} dataset predicts a Pok\'{e}mon's primary type (\texttt{Type 1}) from its name and battle statistics.

\paragraph{Tabula-8B Format:}
{\small
\begin{verbatim}
<|begin_of_text|>Predict the value of Type 1: ||Steel||Normal||
Rock||...||Flying|| The # is 718. The Name is Zygarde50% Forme.
The Type 2 is Ground. The Total is 600. The HP is 108. The Attack
is 100. The Defense is 121. The Sp Atk is 81. The Sp Def is 95.
The Speed is 95. The Generation is 6. The Legendary is True.
What is the value of Type 1? ||Steel||Normal||...||Flying||
<|endinput|>Dragon<|endcompletion|>
[3 more few-shot examples]
Predict the value of Type 1: ... The # is 319. The Name is
Sharpedo. ...<|endinput|>
\end{verbatim}
}

\paragraph{Alpaca Format:}
{\small
\begin{verbatim}
Below is an instruction that describes a task. Write a response
that appropriately completes the request.

### Instruction:
Predict the correct value based on the input.

### Input:

Example 1:
Predict the value of Type 1: ||Steel||Normal||Rock||Grass||Bug||
Ground||Fairy||Water||Electric||Ghost||Dragon||Ice||Dark||Fire||
Fighting||Psychic||Poison||Flying|| The # is 718. The Name is
Zygarde50% Forme. The Type 2 is Ground. The Total is 600. The HP
is 108. The Attack is 100. The Defense is 121. The Sp Atk is 81.
The Sp Def is 95. The Speed is 95. The Generation is 6. The
Legendary is True. What is the value of Type 1? ||Steel||Normal||
Rock||Grass||Bug||Ground||Fairy||Water||Electric||Ghost||Dragon||
Ice||Dark||Fire||Fighting||Psychic||Poison||Flying||
Response: Dragon

[Examples 2-4 follow same format]

Now complete the following:
Predict the value of Type 1: ||Steel||Normal||Rock||...||Flying||
The # is 319. The Name is Sharpedo. The Type 2 is Dark. The Total
is 460. The HP is 70. The Attack is 120. The Defense is 40. The
Sp Atk is 95. The Sp Def is 40. The Speed is 95. The Generation
is 3. The Legendary is False. What is the value of Type 1?
||Steel||Normal||Rock||...||Flying||

### Response:
\end{verbatim}
}
\subsubsection{Binary Classification: US Women's Labor Force Participation}

The \texttt{us-womens-labor-force-participation} dataset predicts labor force participation (\texttt{lfp}) from demographic features.

\paragraph{Tabula-8B Format:}
{\small
\begin{verbatim}
<|begin_of_text|>Predict the value of lfp: ||1||0|| The Unnamed:
0 is 235. The hours is 1640. The kids5 is 0. The kids618 is 1.
The age is 38. The educ is 12. The wage is 3.658. The repwage is
3.25. The hushrs is 2040. The husage is 46. The huseduc is 12.
The huswage is 3.186. The faminc is 12500. The mtr is 0.721. The
motheduc is 7. The fatheduc is 7. The unem is 7.5. The city is 0.
The exper is 19. The nwifeinc is 6.5. The wifecoll is FALSE. The
huscoll is FALSE. What is the value of lfp? ||1||0||<|endinput|>1
<|endcompletion|>...[3 more examples]...<|endinput|>
\end{verbatim}
}

\paragraph{Alpaca Format:}
{\small
\begin{verbatim}
Below is an instruction that describes a task. Write a response
that appropriately completes the request.

### Instruction:
Predict the correct value based on the input.

### Input:

Example 1:
Predict the value of lfp: ||1||0|| The Unnamed: 0 is 235. The
hours is 1640. The kids5 is 0. The kids618 is 1. The age is 38.
The educ is 12. The wage is 3.658. The repwage is 3.25. The
hushrs is 2040. The husage is 46. The huseduc is 12. The huswage
is 3.186. The faminc is 12500. The mtr is 0.721. The motheduc is
7. The fatheduc is 7. The unem is 7.5. The city is 0. The exper
is 19. The nwifeinc is 6.5. The wifecoll is FALSE. The huscoll is
FALSE. What is the value of lfp? ||1||0||
Response: 1

[Examples 2-4 omitted for brevity]

Now complete the following:
Predict the value of lfp: ||1||0|| The Unnamed: 0 is 728. The
hours is 0. The kids5 is 0. The kids618 is 3. The age is 41. The
educ is 12. The wage is 0.0. The repwage is 0.0. The hushrs is
2450. The husage is 48. The huseduc is 15. The huswage is 26.53.
The faminc is 77000. The mtr is 0.442. The motheduc is 7. The
fatheduc is 14. The unem is 14.0. The city is 1. The exper is 8.
The nwifeinc is 77.0. The wifecoll is FALSE. The huscoll is TRUE.
What is the value of lfp? ||1||0||

### Response:
\end{verbatim}
}

%Note that the test instance (row 728) corresponds exactly to the contamination example shown in Figure~\ref{fig:contamination}(a).

\subsubsection{Quartile Classification: Bitcoin Price Prediction}

The \texttt{bitcoin-price-2014-2023} dataset predicts the closing price discretized into quartile bins, illustrating the quartile classification format.

\paragraph{Tabula-8B Format:}
{\small
\begin{verbatim}
<|begin_of_text|>Predict the value of Close: ||between 7697.924072
and 20297.0288085||less than 764.11325075||between 764.11325075
and 7697.924072||greater than 20297.0288085|| The Date is
2023-07-09. The Open is 30291.611328125. The High is
30427.58984375. The Low is 30085.591796875. The Adj Close is
30171.234375. The Volume is 7903327744. What is the value of
Close? ||between 7697.924072 and 20297.0288085||less than
764.11325075||between 764.11325075 and 7697.924072||greater than
20297.0288085||<|endinput|>greater than 20297.0288085
<|endcompletion|>...[3 more examples]...<|endinput|>
\end{verbatim}
}

\paragraph{Alpaca Format:}
{\small
\begin{verbatim}
Below is an instruction that describes a task. Write a response
that appropriately completes the request.

### Instruction:
Predict the correct value based on the input.

### Input:

Example 1:
Predict the value of Close: ||between 7697.924072 and
20297.0288085||less than 764.11325075||between 764.11325075 and
7697.924072||greater than 20297.0288085|| The Date is 2023-07-09.
The Open is 30291.611328125. The High is 30427.58984375. The Low
is 30085.591796875. The Adj Close is 30171.234375. The Volume is
7903327744. What is the value of Close? ||between 7697.924072 and
20297.0288085||less than 764.11325075||between 764.11325075 and
7697.924072||greater than 20297.0288085||
Response: greater than 20297.0288085

[Examples 2-4 omitted for brevity]

Now complete the following:
Predict the value of Close: ||between 7697.924072 and
20297.0288085||less than 764.11325075||between 764.11325075 and
7697.924072||greater than 20297.0288085|| The Date is 2019-05-27.
The Open is 8674.072265625. The High is 8907.1748046875. The Low
is 8668.705078125. The Adj Close is 8805.7783203125. The Volume
is 27949840384. What is the value of Close? ||between 7697.924072
and 20297.0288085||less than 764.11325075||between 764.11325075
and 7697.924072||greater than 20297.0288085||

### Response:
\end{verbatim}
}

\subsection{Alpaca+Q Training}

For quartile classification tasks, the standard Alpaca model performed poorly due to unfamiliarity with the quartile bin output format.
These tasks require predicting labels such as ``less than 25.3'' or ``between 25.3 and 50.1,'' a format not present in the general Alpaca training data.

To isolate format learning from tabular knowledge, we created Alpaca+Q by augmenting the Alpaca training data with 10K quartile-format examples.
These examples were drawn from 20 regression datasets that we held out from all evaluation analyses reported in Section~\ref{sec:format_learning}.
For each held-out dataset, we discretized the continuous target into quartile bins and formatted examples using the same serialization as Tabula-8B evaluation.
This ensures that any performance gains from Alpaca+Q reflect familiarity with the quartile output format rather than exposure to evaluation data.

The 20 held-out datasets used for Alpaca+Q training are:
\begin{enumerate}[noitemsep]
\item \texttt{aakashjoshi123/exercise-and-fitness-metrics-dataset}
\item \texttt{altruistdelhite04/gold-price-data}
\item \texttt{amirhosseinmirzaie/countries-life-expectancy}
\item \texttt{arnabchaki/data-science-salaries-2023}
\item \texttt{arslanr369/roblox-stock-pricing-2021-2023}
\item \texttt{awaiskaggler/insurance-csv}
\item \texttt{dsfelix/us-stores-sales}
\item \texttt{gyanprakashkushwaha/laptop-price-prediction-cleaned-dataset}
\item \texttt{harshitshankhdhar/imdb-dataset-of-top-1000-movies-and-tv-shows}
\item \texttt{hemanthhari/psycological-effects-of-covid}
\item \texttt{iamsumat/spotify-top-2000s-mega-dataset}
\item \texttt{mirichoi0218/insurance}
\item \texttt{noordeen/insurance-premium-prediction}
\item \texttt{oles04/top-leagues-player}
\item \texttt{prevek18/ames-housing-dataset}
\item \texttt{shreyapurohit/anime-data}
\item \texttt{suraj520/dairy-goods-sales-dataset}
\item \texttt{teertha/ushealthinsurancedataset}
\item \texttt{warcoder/earthquake-dataset}
\item \texttt{whenamancodes/students-performance-in-exams}
\end{enumerate}

\subsection{Evaluation Protocol}

For all models (Base Llama, Alpaca, Alpaca+Q, and Tabula-8B), we used the same evaluation protocol: few-shot prompting with serialized table rows, followed by extracting the model's predicted class from the generated completion.
We used greedy decoding (temperature = 0) to ensure reproducibility.
All comparisons use lift over the majority-class baseline as the primary metric, which controls for class imbalance and enables fair comparison across datasets with varying class distributions.

\section{Contamination Detection Methodology}
\label{app:contamination_methodology}

\subsection{Challenges for Tabular Contamination Detection}

Data contamination detection in tabular settings poses unique challenges not present in text or image domains.
Standard approaches based on exact string matching or $n$-gram overlap~\cite{brown2020language} are insufficient for several reasons:

\begin{enumerate}[leftmargin=*,noitemsep]
\item \textbf{Permutation invariance.} Tabular data is row-permutation and column-permutation invariant. The same dataset can appear with columns in different orders or rows shuffled, defeating exact-match detection.

\item \textbf{Column name variations.} The same underlying data frequently appears across web sources with different column names. For example, we found the \texttt{us-womens-labor} dataset with columns named \texttt{kids618} in the evaluation set but \texttt{k618} in T4, and \texttt{nwifeinc} versus \texttt{inc} for the same field.

\item \textbf{Multi-source duplication.} Popular datasets (stock prices, well-known Kaggle competitions) are replicated across many websites, often with slight schema variations. A single evaluation record may appear in dozens of T4 tables under different presentations.

\item \textbf{Task-level leakage.} Even when specific evaluation records are absent from training, the \emph{task} may be solvable from training data. For instance, predicting day-of-week from a date requires no dataset-specific knowledge if the model has memorized date-day associations from other sources.
\end{enumerate}

\subsection{Search Procedure}

We focused our contamination analysis on the top-performing datasets (Table~\ref{tab:top_datasets}), where contamination would be most consequential for claims of strong tabular reasoning.
The T4 corpus comprises approximately 150 zip archives containing millions of parquet files.
Due to computational constraints, we conducted targeted searches rather than exhaustive corpus-wide matching.

For each dataset examined, we employed the following search strategies:

\paragraph{Distinctive identifier search.}
For datasets with unique identifiers (e.g., Pokemon names, course IDs, stock ticker symbols), we searched T4 for these identifiers to locate potentially matching tables.
This approach reduces false positives from common values while efficiently locating relevant training data.

\paragraph{Row-level value matching.}
We searched for combinations of field values that would be unlikely to co-occur by chance.
For the \texttt{us-womens-labor} dataset, we searched for records containing the distinctive column names (\texttt{kids5}, \texttt{hushrs}, \texttt{nwifeinc}) and verified field-level matches for identified records.

\paragraph{Task-level association search.}
For tasks reducible to factual associations (e.g., date $\rightarrow$ day-of-week), we searched T4 for the underlying associations rather than specific evaluation records.
For the \texttt{peloton-data} dataset, we queried for records containing both a sample date (``2021-11-30'') and its corresponding day (``Tuesday''), finding 844 matches from unrelated tables.

% \subsection{Contamination Types Identified}

% Our analysis revealed three distinct contamination patterns:

% \paragraph{Complete train-test overlap.}
% All evaluation rows appear in training data with labels exposed.
% Example: \texttt{us-womens-labor} (753/753 rows in T4 with target labels).

% \paragraph{Direct contamination with label exposure.}
% Evaluation records appear across multiple T4 tables, often with the prediction target explicitly included.
% Example: Bitcoin price records appear in up to 6 separate T4 chunks, each containing the \texttt{Close} value the model is asked to predict.

% \paragraph{Task leakage.}
% The evaluation dataset passes row-level deduplication, but the task remains solvable via memorized associations from unrelated training data.
% Example: \texttt{peloton-data} contains no exact matches in T4, but 844 T4 records encode the date-day mapping required to solve the task.

\subsection{Limitations}

Our contamination analysis has several limitations:

\begin{enumerate}[leftmargin=*,noitemsep]
\item \textbf{Non-exhaustive search.} We searched for specific patterns rather than conducting corpus-wide exact matching. Our analysis can demonstrate the \emph{presence} of contamination but not definitively establish its absence.

\item \textbf{Focus on top performers.} We prioritized datasets where contamination would most affect performance claims. Lower-performing datasets were not systematically examined.

\item \textbf{Detection methodology limitations.} Tabular contamination detection lacks established best practices. More sophisticated approaches (e.g., fuzzy matching, semantic similarity) may reveal additional contamination we did not detect.
\end{enumerate}

These limitations suggest our findings represent a lower bound on the true extent of contamination in the benchmark.

\section{Extended Task-heterogeneity Results}
\begin{table}[thb]
\centering
\caption{Datasets with negative Cohen's Kappa, indicating classifier agreement worse than chance. Metrics include raw accuracy (Acc.), majority-class baseline accuracy (Maj.), lift over majority-class baseline (Lift) and Cohen's Kappa ($\kappa$).}
\label{tab:negative_kappa}
\vspace{0.5em}
\begin{tabular}{l|l|cccc}
\toprule
\textsc{Dataset} & \textsc{Type} & \textsc{Acc.} & \textsc{Maj.} & \textsc{Lift} & $\kappa$ \\
\midrule
hawkingcr/airbnb-for-boston-with-fraud-detection &  Binary &  0.749 & 0.792 & -0.043 & -0.051 \\
tejashvi14/employee-future-prediction &  Binary &  0.544 & 0.650 & -0.106 & -0.047 \\
adityakadiwal/water-potability &  Binary &  0.472 & 0.610 & -0.139 & -0.034 \\
buntyshah/auto-insurance-claims-data &  Binary &  0.655 & 0.753 & -0.098 & -0.030 \\
vstacknocopyright/blood-transfusion-service-center-data &  Binary &  0.553 & 0.762 & -0.209 & -0.029 \\
ahsan81/food-ordering-and-delivery-app-dataset &  Categorical &  0.246 & 0.400 & -0.154 & -0.021 \\
arnavsmayan/netflix-userbase-dataset &  Categorical &  0.333 & 0.422 & -0.089 & -0.019 \\
receplyasolu/6k-weather-labeled-spotify-songs &  Categorical &  0.166 & 0.280 & -0.114 & -0.018 \\
arnavsmayan/vehicle-manufacturing-dataset &  Categorical &  0.054 & 0.111 & -0.058 & -0.009 \\
kumargh/pimaindiansdiabetescsv &  Categorical &  0.111 & 0.176 & -0.065 & -0.005 \\
hansrobertson/american-companies-profits-and-benefits-from-ai &  Quartile &  0.246 & 0.253 & -0.007 & -0.005 \\
jillanisofttech/brain-stroke-dataset &  Binary &  0.958 & 0.959 & -0.001 & -0.002 \\
\bottomrule
\end{tabular}
\end{table}

\begin{table}[htb]
\centering
\caption{Performance by task type, measured as lift over majority-class baseline (mean and median) and and percentage of datasets performing at or below the baseline ($\le$ Baseline)}
\label{tab:task_type_performance}
\vspace{0.5em}
\begin{tabular}{l|c|cc|c}
\toprule
\textsc{Task Type} & \textsc{N} & \textsc{Mean Lift}& \textsc{Median Lift} & \textsc{$\leq$ Maj. Baseline (\%)} \\
\midrule
Quartile & 46 & +35.6 & +32.9 & 10.9 \\
Categorical & 55 & +10.9 & $-$0.3 & 54.5\\
Binary & 64 & +2.7 & +2.5 & 46.9\\
\bottomrule
\end{tabular}
\end{table}

\begin{table}[htb]
\centering
\caption{Pairwise $t$-tests comparing lift over majority-class baseline across task types. All comparisons are statistically significant at 0.05 significance threshold.}
\label{tab:pairwise_ttests}
\vspace{0.5em}
\begin{tabular}{l|c|c}
\toprule
\textsc{Comparison} & \textsc{$t$-statistic} & \textsc{$p$-value} \\
\midrule
Quartile vs.\ Binary & 8.73 & $3.49 \times 10^{-14}$ \\
Quartile vs.\ Categorical & 4.62 & $1.52 \times 10^{-5}$ \\
Categorical vs.\ Binary & 2.25 & $2.65 \times 10^{-2}$ \\
\bottomrule
\end{tabular}
\end{table}

\begin{figure}[b]
\centering
\includegraphics[width=0.9\linewidth]{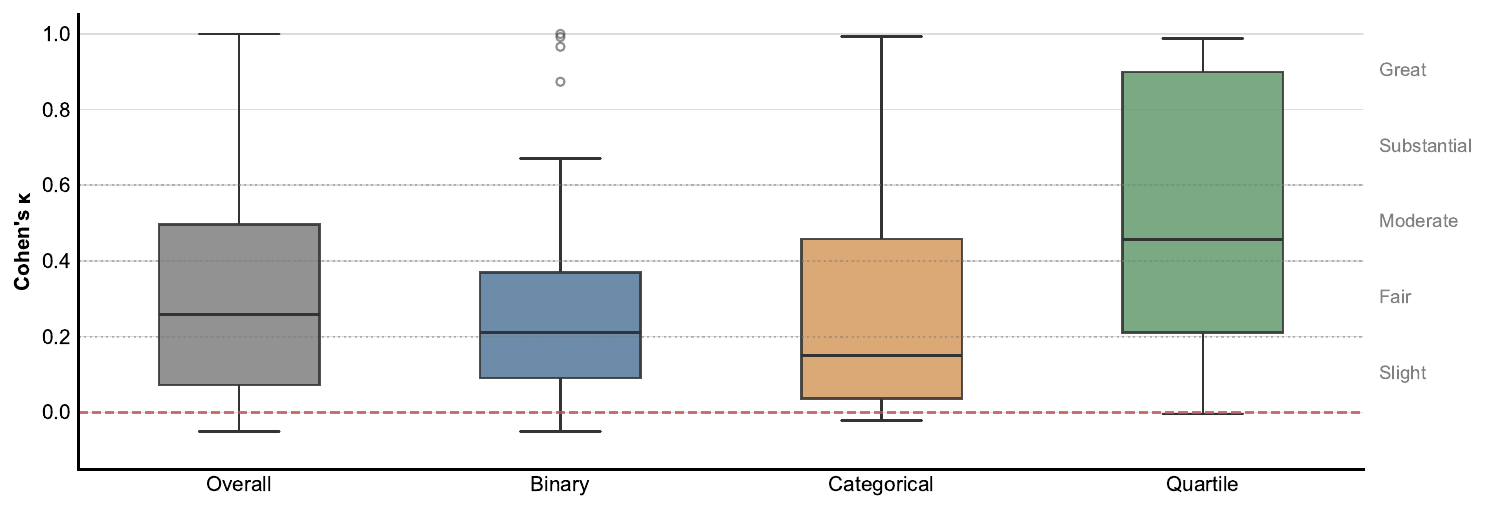}
\caption{Cohen's $\kappa$ distribution by task type. Horizontal lines indicate standard interpretation thresholds.}
\label{fig:kappa_distribution}
\end{figure}

\section{Quartile Classification May Be Artificially Easy}
\label{app:quartile}

Quartile classification tasks discretize continuous targets into four bins based on quartile boundaries. Examining this task format reveals several properties that may inflate performance independent of genuine tabular reasoning:

\begin{enumerate}[leftmargin=*,noitemsep]
     \item \textbf{Numeric leakage.} quartile classification tasks in this benchmark predict a continuous target (e.g., stock closing price) discretized into quartile bins. However, the input features often include highly correlated values. For example, predicting ``Close'' price when ``Open,'' ``High,'' ``Low,'' and ``Adj Close'' are provided as features creates trivial numeric shortcuts---if Open $= 603$, the model can infer Close $> 509$ without learning any generalizable pattern.
    
    \item \textbf{Balanced by construction.} Quartile bins guarantee 25\% of samples per class \emph{by definition}. This is a consequence of how quartiles are computed, not a property of the underlying data (which can be quite skewed or concentrated). In contrast, binary and categorical tasks exhibit natural class imbalance, making majority-class prediction a stronger baseline.
    
    \item \textbf{Ordinal structure.} The four quartile bins (e.g., ``less than X,'' ``between X and Y,'' ``greater than Z'') have inherent ordinal relationships that language models may exploit through numeric reasoning capabilities acquired during pretraining~\cite{shah2023numeric}, rather than learning tabular-specific representations.
\end{enumerate}

\clearpage

\section{Extended Contamination Examples}
\label{app:contamination_examples}

\paragraph{US Women's Labor Force Participation: Extended Evidence.}
As described in Section~\ref{sec:data_leakage}, all 753 test rows from this dataset appear in T4 with matching labels. Table~\ref{tab:lfp_matches} provides the first 20 matched rows. Columns show the T4 field names (\texttt{k5}, \texttt{k618}, \texttt{inc}); corresponding test fields use longer names (\texttt{kids5}, \texttt{kids618}, \texttt{nwifeinc}). Test labels are encoded as 1/0; T4 labels as yes/no. All 753 rows exhibit exact value and label matches.

\begin{table}[h]
\centering
\scriptsize
\begin{tabular}{rrrrrrcc}
\toprule
Row & age & k5 & k618 & inc & Test Label & T4 Label & Match \\
\midrule
1 & 32 & 1 & 0 & 10.9 & 1 & yes & \checkmark \\
2 & 30 & 0 & 2 & 19.5 & 1 & yes & \checkmark \\
3 & 35 & 1 & 3 & 12.0 & 1 & yes & \checkmark \\
4 & 34 & 0 & 3 & 6.8 & 1 & yes & \checkmark \\
5 & 31 & 1 & 2 & 20.1 & 1 & yes & \checkmark \\
6 & 54 & 0 & 0 & 9.9 & 1 & yes & \checkmark \\
7 & 37 & 0 & 2 & 9.2 & 1 & yes & \checkmark \\
8 & 54 & 0 & 0 & 10.9 & 1 & yes & \checkmark \\
9 & 48 & 0 & 2 & 17.3 & 1 & yes & \checkmark \\
10 & 39 & 0 & 2 & 12.9 & 1 & yes & \checkmark \\
11 & 33 & 0 & 1 & 24.3 & 1 & yes & \checkmark \\
12 & 42 & 0 & 1 & 19.7 & 1 & yes & \checkmark \\
13 & 30 & 1 & 2 & 15.0 & 1 & yes & \checkmark \\
14 & 43 & 0 & 2 & 14.6 & 1 & yes & \checkmark \\
15 & 43 & 0 & 1 & 24.6 & 1 & yes & \checkmark \\
16 & 35 & 0 & 3 & 17.5 & 1 & yes & \checkmark \\
17 & 43 & 0 & 2 & 14.1 & 1 & yes & \checkmark \\
18 & 39 & 0 & 5 & 15.8 & 1 & yes & \checkmark \\
19 & 45 & 0 & 0 & 14.1 & 1 & yes & \checkmark \\
20 & 35 & 0 & 4 & 10.3 & 1 & yes & \checkmark \\
\midrule
\multicolumn{8}{c}{\textit{... 733 additional matching rows omitted ...}} \\
\bottomrule
\end{tabular}
\caption{First 20 rows of contamination evidence for \texttt{us-womens-labor-force-participation}. All 753 test rows appear in T4 (chunk-0098) with matching labels. Column names differ slightly (e.g., \texttt{kids618} $\rightarrow$ \texttt{k618}, \texttt{nwifeinc} $\rightarrow$ \texttt{inc}).}
\label{tab:lfp_matches}
\end{table}

\begin{figure}[t]
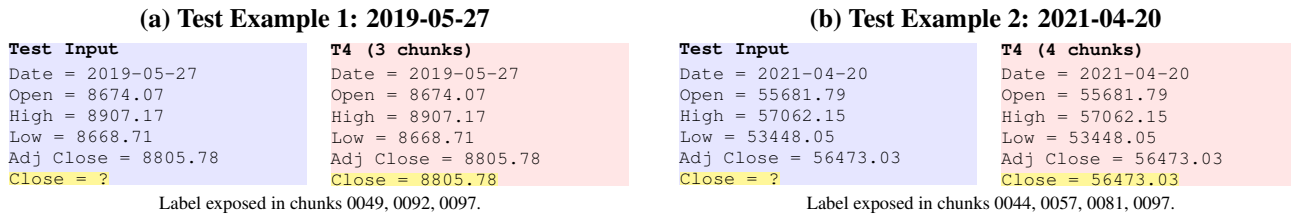

\centering
\textbf{Complete Overlap: bitcoin-price-2014-2023}
\vspace{0.5em}

\begin{subfigure}[t]{0.48\textwidth}
\centering
\textbf{(a) Test Example 1: 2019-05-27}
\vspace{0.3em}

\begin{minipage}[t]{0.48\textwidth}
\colorbox{blue!10}{%
\parbox{\dimexpr\linewidth-2\fboxsep}{%
\scriptsize\ttfamily
\textbf{Test Input}\\[0.2em]
Date = 2019-05-27\\
Open = 8674.07\\
High = 8907.17\\
Low = 8668.71\\
Adj Close = 8805.78\\
\colorbox{yellow!50}{Close = ?}
}}
\end{minipage}%
\hfill%
\begin{minipage}[t]{0.48\textwidth}
\colorbox{red!10}{%
\parbox{\dimexpr\linewidth-2\fboxsep}{%
\scriptsize\ttfamily
\textbf{T4 (3 chunks)}\\[0.2em]
Date = 2019-05-27\\
Open = 8674.07\\
High = 8907.17\\
Low = 8668.71\\
Adj Close = 8805.78\\
\colorbox{yellow!50}{Close = 8805.78}
}}
\end{minipage}

\vspace{0.3em}
\scriptsize Label exposed in chunks 0049, 0092, 0097.
\end{subfigure}%
\hfill%
\begin{subfigure}[t]{0.48\textwidth}
\centering
\textbf{(b) Test Example 2: 2021-04-20}
\vspace{0.3em}

\begin{minipage}[t]{0.48\textwidth}
\colorbox{blue!10}{%
\parbox{\dimexpr\linewidth-2\fboxsep}{%
\scriptsize\ttfamily
\textbf{Test Input}\\[0.2em]
Date = 2021-04-20\\
Open = 55681.79\\
High = 57062.15\\
Low = 53448.05\\
Adj Close = 56473.03\\
\colorbox{yellow!50}{Close = ?}
}}
\end{minipage}%
\hfill%
\begin{minipage}[t]{0.48\textwidth}
\colorbox{red!10}{%
\parbox{\dimexpr\linewidth-2\fboxsep}{%
\scriptsize\ttfamily
\textbf{T4 (4 chunks)}\\[0.2em]
Date = 2021-04-20\\
Open = 55681.79\\
High = 57062.15\\
Low = 53448.05\\
Adj Close = 56473.03\\
\colorbox{yellow!50}{Close = 56473.03}
}}
\end{minipage}

\vspace{0.3em}
\scriptsize Label exposed in chunks 0044, 0057, 0081, 0097.
\end{subfigure}

\vspace{0.5em}
%\small In both cases, \texttt{Adj Close} $\approx$ \texttt{Close} also enables trivial inference without memorization.

\caption{Contamination in Bitcoin price prediction. Test examples appear verbatim in multiple T4 chunks with target labels exposed. Additionally, \texttt{Adj Close} $\approx$ \texttt{Close} provides a numeric shortcut.}
\label{fig:bitcoin_contamination}
\end{figure}

\paragraph{Bitcoin Price: Extended Match Details.}
Figure~\ref{fig:bitcoin_contamination} provides detailed evidence of contamination in the Bitcoin price prediction task. Two representative test examples (2019-05-27 and 2021-04-20) each appear in multiple T4 chunks with the target \texttt{Close} value fully exposed. The first example appears in three chunks; the second in four. This demonstrates that individual Bitcoin price records are replicated extensively throughout T4, with the model encountering the same labeled examples multiple times during training.

\begin{figure}[t]
\centering
\textbf{Financial Time Series Contamination}
\vspace{0.5em}

\begin{subfigure}[t]{0.48\textwidth}
\centering
\textbf{(a) Netflix Stock: 2020-03-19}
\vspace{0.3em}

\begin{minipage}[t]{0.48\textwidth}
\colorbox{blue!10}{%
\parbox{\dimexpr\linewidth-2\fboxsep}{%
\scriptsize\ttfamily
\textbf{Test Input}\\[0.2em]
Date = 2020-03-19\\
Open = 324.33\\
High = 348.51\\
Low = 316.82\\
Adj Close = 332.03\\
\colorbox{yellow!50}{Close = ?}
}}
\end{minipage}%
\hfill%
\begin{minipage}[t]{0.48\textwidth}
\colorbox{red!10}{%
\parbox{\dimexpr\linewidth-2\fboxsep}{%
\scriptsize\ttfamily
\textbf{T4: chunk-0045}\\[0.2em]
Open = 324.33\\
High = 348.51\\
Low = 316.82\\
Adj Close = 332.03\\
\colorbox{yellow!50}{Close = 332.03}
}}
\end{minipage}
\end{subfigure}%
\hfill%
\begin{subfigure}[t]{0.48\textwidth}
\centering
\textbf{(b) Bitcoin-USD: 2019-04-30}
\vspace{0.3em}

\begin{minipage}[t]{0.48\textwidth}
\colorbox{blue!10}{%
\parbox{\dimexpr\linewidth-2\fboxsep}{%
\scriptsize\ttfamily
\textbf{Test Input}\\[0.2em]
Date = 2019-04-30\\
Open = 5247.73\\
High = 5363.26\\
Low = 5224.19\\
Adj Close = 5350.73\\
\colorbox{yellow!50}{Close = ?}
}}
\end{minipage}%
\hfill%
\begin{minipage}[t]{0.48\textwidth}
\colorbox{red!10}{%
\parbox{\dimexpr\linewidth-2\fboxsep}{%
\scriptsize\ttfamily
\textbf{T4: chunk-0097}\\[0.2em]
Open = 5247.73\\
High = 5363.26\\
Low = 5224.19\\
Adj Close = 5350.73\\
\colorbox{yellow!50}{Close = 5350.73}
}}
\end{minipage}
\end{subfigure}

\vspace{1em}

\begin{subfigure}[t]{0.48\textwidth}
\centering
\textbf{(c) Yahoo Stock: 2018-09-11}
\vspace{0.3em}

\begin{minipage}[t]{0.48\textwidth}
\colorbox{blue!10}{%
\parbox{\dimexpr\linewidth-2\fboxsep}{%
\scriptsize\ttfamily
\textbf{Test Input}\\[0.2em]
Date = 2018-09-11\\
Open = 2871.57\\
High = 2892.52\\
Low = 2866.78\\
Adj Close = 2887.89\\
\colorbox{yellow!50}{Close = ?}
}}
\end{minipage}%
\hfill%
\begin{minipage}[t]{0.48\textwidth}
\colorbox{red!10}{%
\parbox{\dimexpr\linewidth-2\fboxsep}{%
\scriptsize\ttfamily
\textbf{T4: chunk-0054}\\[0.2em]
Open = 2871.57\\
High = 2892.52\\
Low = 2866.78\\
Adj Close = 2887.89\\
\colorbox{yellow!50}{Close = 2887.89}
}}
\end{minipage}
\end{subfigure}%
\hfill%
\begin{subfigure}[t]{0.48\textwidth}
\centering
\textbf{(d) Tesla Stock: 2017-06-26}
\vspace{0.3em}

\begin{minipage}[t]{0.48\textwidth}
\colorbox{blue!10}{%
\parbox{\dimexpr\linewidth-2\fboxsep}{%
\scriptsize\ttfamily
\textbf{Test Input}\\[0.2em]
Date = 2017-06-26\\
Open = 386.69\\
High = 386.95\\
Low = 373.10\\
Adj Close = 377.49\\
\colorbox{yellow!50}{Close = ?}
}}
\end{minipage}%
\hfill%
\begin{minipage}[t]{0.48\textwidth}
\colorbox{red!10}{%
\parbox{\dimexpr\linewidth-2\fboxsep}{%
\scriptsize\ttfamily
\textbf{T4: chunk-0017}\\[0.2em]
Open = 386.69\\
High = 386.95\\
Low = 373.10\\
Adj Close = 377.49\\
\colorbox{yellow!50}{Close = 377.49}
}}
\end{minipage}
\end{subfigure}

\caption{Contamination in financial time series datasets. Test examples appear in T4 with target \texttt{Close} values exposed. These datasets rank among the top 10 performers in the Tabula-8B evaluation.}
\label{fig:financial_contamination}
\end{figure}

\paragraph{Other Financial Time Series.}
The contamination pattern extends beyond Bitcoin to other top-performing financial datasets. Figure~\ref{fig:financial_contamination} shows representative examples from Netflix, Bitcoin-USD, Yahoo, and Tesla stock prediction tasks. Each exhibits the same pattern of test records appearing in T4 with target labels fully exposed. These five financial time series collectively account for half of the top 10 performers in Table~\ref{tab:top_datasets}, suggesting that the strong benchmark results on financial forecasting tasks are driven by data memorization rather than genuine predictive capability.

\paragraph{Peloton: Extended Match Details.}
Table~\ref{tab:peloton_date_matches} shows the distribution of the date-day association (2021-11-30 $\rightarrow$ Tuesday) across T4. All 844 records contain both the date and its corresponding day label, spanning power grids (chunk-0086), transit systems (chunks-0015, 0115), telecom networks (chunk-0144), and 21 other unrelated domains. The Peloton workout dataset itself does not appear in T4, yet the task of predicting day-of-week from date is trivially solvable via memorization of this widely-attested association across hundreds of timestamped operational datasets.

\begin{table}[t]
\centering
\caption{Task leakage evidence: Distribution of date-day associations in T4 training corpus. The mapping 2021-11-30 $\rightarrow$ Tuesday appears 844 times across 24 different T4 chunks spanning diverse domains.}
\label{tab:peloton_date_matches}
\small
\begin{tabular}{@{}lrp{0.58\linewidth}@{}}
\toprule
\textbf{Dataset Domain} & \textbf{Matches} & \textbf{Representative Fields/Context} \\
\midrule
Power grid monitoring & 637 & Block, Time, Avg. Frequency, SG(MW), AG(MW), Unit\_Running \\
Bus transit schedules & 91 & Transit Date, Block \#, Badge \#, Bus Stop ID, Latitude, Longitude \\
Transit delays & 38 & Date, Line (500--505), Location, Incident type, Min Delay, Vehicle \\
4G network metrics & 24 & Period, Data Accessibility SR, Voice DCR, ERAB Attempts, Hour \\
Prayer times & 1 & Date, Fajr, Sunrise, Zuhr, Asr, Maghrib, Isha times \\
Meal delivery platform & 1 & TotalSupply, MealsSaved, PayingUsers, SupplyingStores \\
Other timestamped data & 52 & Various operational/monitoring systems across 18 chunks \\
\midrule
\textbf{Total across 24 chunks} & \textbf{844} & \\
\bottomrule
\end{tabular}
\end{table}

\clearpage
\section{Alpaca Extended Results} 

\begin{table}[H]
\caption{Accuracy (\%) across for Tabula-8B, Base Llama, and Alpaca, for \texttt{bundesliga-seasons} and \texttt{udemy-courses}.}
\label{tab:format_results}
\begin{center}
\begin{small}
\begin{sc}

\begin{tabular}{l|l|cccc}
\toprule
\textbf{Dataset} & \textbf{Task} & \textbf{Base} & \textbf{Alpaca} & \textbf{Tabula} \\
\midrule
bundesliga & Binary & 94.5 & 97.2 & 100.0 \\
udemy & Categ. & 70.6 & 94.2 & 99.5 \\
% pokemon & Categ. & XX.X & 87.9 & 99.0 \\
% bitcoin & Quart. & XX.X & 32.6 & 99.1 \\
\bottomrule
\end{tabular}
\end{sc}
\end{small}
\end{center}
\vskip -0.2in
\end{table}

\begin{table}[h!]
\caption{Classification performance before and after excluding datasets with 
documented contamination: \texttt{pokemon} ($\times2$), \texttt{peloton-data}, 
\texttt{us-womens-labor}, and \texttt{udemy-courses}\footnotemark. Lift over majority-class baseline shown as mean and median across datasets. Gap indicates difference between Alpaca and Tabula-8B mean lift.}
\label{tab:contamination_exclusion}
\begin{center}
\begin{small}
\begin{sc}
\begin{tabular}{l|cc|cc|cc|c}
\toprule
& \multicolumn{2}{c|}{\textbf{Tabula-8B}} & \multicolumn{2}{c|}{\textbf{Alpaca}} & \multicolumn{2}{c|}{\textbf{Base Llama}} & \textbf{Gap} \\
\textbf{Dataset Set} & \textbf{Mean} & \textbf{Median} & \textbf{Mean} & \textbf{Median} & \textbf{Mean} & \textbf{Median} & \textbf{(Mean)} \\
\midrule
All Classification (N=119) & +6.5 & +0.0 & +1.6 & -1.5 & -9.4 & -5.4 & +4.9 \\
Excl. Contaminated (N=114) & +4.3 & -0.1 & +0.3 & -1.7 & -10.4 & -5.8 & +4.1 \\
\midrule
Change & -2.2 & -0.1 & -1.3 & -0.2 & -1.0 & -0.4 & -0.8 \\
\bottomrule
\end{tabular}
\end{sc}
\end{small}
\end{center}
\vskip -0.2in
\end{table}

\footnotetext{The \texttt{udemy-courses} task classifies courses into four 
subjects based on \texttt{course\_title}; likely a text classification task 
solvable through pretrained LLMs.}

% Ultra-compact table matching paper format
% Required: \usepackage{longtable,booktabs}

{\small % Use small font for entire table
\setlength{\tabcolsep}{4pt} % Reduce column separation
\begin{longtable}{@{}p{7.5cm} |l |c c c c c@{}}
\caption{Accuracy scores for Tabula-8B, Llama-3-8B, and Alpaca variants across 165 tabular datasets. Alpaca is Llama-3-8B fine-tuned on the Alpaca instruction dataset, while Alpaca+Q is further trained with quartile regression data. Majority baseline predicts the most frequent class. Empty cells indicate tasks used to train Alpaca+Q.}
\label{tab:model_comparison} \\
\toprule
\textbf{Dataset} & \textbf{Type} & \textbf{Majority} & \textbf{Tabula-8B} & \textbf{Llama-3-8B} & \textbf{Alpaca} & \textbf{Alpaca+Q} \\
\midrule
\endfirsthead
\toprule
\textbf{Dataset} & \textbf{Type} & \textbf{Majority} & \textbf{Tabula-8B} & \textbf{Llama-3-8B} & \textbf{Alpaca} & \textbf{Alpaca+Q} \\
\midrule
\endhead
\bottomrule
\endfoot
\bottomrule
\endlastfoot
aakashjoshi123/exercise-and-fitness-metrics-dataset & Quartile & 0.280 & 0.577 & 0.064 & 0.299 &  \\
aakashjoshi123/spotify-top-hits-data & Categorical & 0.757 & 0.721 & 0.708 & 0.720 & 0.729 \\
abcsds/pokemon & Categorical & 0.140 & 0.990 & 0.677 & 0.882 & 0.968 \\
adityakadiwal/water-potability & Binary & 0.610 & 0.472 & 0.498 & 0.519 & 0.557 \\
agirlcoding/all-space-missions-from-1957 & Categorical & 0.903 & 0.904 & 0.833 & 0.876 & 0.899 \\
ahsan81/food-ordering-and-delivery-app-dataset & Categorical & 0.400 & 0.246 & 0.248 & 0.263 & 0.275 \\
ahsan81/superstore-marketing-campaign-dataset & Binary & 0.849 & 0.730 & 0.663 & 0.778 & 0.634 \\
akshaydattatraykhare/diabetes-dataset & Binary & 0.651 & 0.707 & 0.674 & 0.694 & 0.727 \\
alexisbcook/pakistan-intellectual-capital & Categorical & 0.650 & 0.649 & 0.216 & 0.281 & 0.024 \\
alirezachahardoli/bank-personal-loan-1 & Binary & 0.899 & 0.843 & 0.879 & 0.865 & 0.847 \\
altruistdelhite04/gold-price-data & Quartile & 0.265 & 0.742 & 0.121 & 0.289 &  \\
amirhosseinmirzaie/countries-life-expectancy & Quartile & 0.266 & 0.756 & 0.273 & 0.353 &  \\
amirhosseinmirzaie/pistachio-types-detection & Binary & 0.575 & 0.668 & 0.417 & 0.519 & 0.590 \\
ananthr1/weather-prediction & Categorical & 0.438 & 0.774 & 0.001 & 0.604 & 0.668 \\
andrewmvd/fetal-health-classification & Categorical & 0.791 & 0.742 & 0.733 & 0.662 & 0.514 \\
andrewmvd/udemy-courses & Categorical & 0.316 & 0.995 & 0.706 & 0.942 & 0.988 \\
arashnic/time-series-forecasting-with-yahoo-stock-price & Quartile & 0.257 & 0.988 & 0.244 & 0.409 & 0.731 \\
arnabchaki/data-science-salaries-2023 & Quartile & 0.273 & 0.947 & 0.001 & 0.556 &  \\
arnabchaki/indian-restaurants-2023 & Quartile & 0.365 & 0.276 & 0.128 & 0.303 & 0.300 \\
arnavsmayan/netflix-userbase-dataset & Categorical & 0.422 & 0.333 & 0.344 & 0.378 & 0.363 \\
arnavsmayan/vehicle-manufacturing-dataset & Categorical & 0.111 & 0.054 & 0.066 & 0.083 & 0.077 \\
arslanr369/bitcoin-price-2014-2023 & Quartile & 0.263 & 0.991 & 0.175 & 0.326 & 0.829 \\
arslanr369/roblox-stock-pricing-2021-2023 & Quartile & 0.250 & 0.969 & 0.012 & 0.355 &  \\
ashishkumarjayswal/diabetes-dataset & Binary & 0.651 & 0.677 & 0.648 & 0.669 & 0.712 \\
ashishkumarjayswal/loanamount-approval & Binary & 0.687 & 0.780 & 0.169 & 0.596 & 0.630 \\
ashishkumarjayswal/movies-updated-data & Categorical & 0.342 & 0.637 & 0.479 & 0.580 & 0.554 \\
atharvaingle/crop-recommendation-dataset & Categorical & 0.052 & 0.223 & 0.000 & 0.057 & 0.101 \\
awaiskaggler/insurance-csv & Quartile & 0.265 & 0.624 & 0.068 & 0.293 &  \\
azminetoushikwasi/-lionel-messi-all-club-goals & Categorical & 0.619 & 0.445 & 0.214 & 0.385 & 0.284 \\
barun2104/telecom-churn & Binary & 0.863 & 0.851 & 0.748 & 0.798 & 0.810 \\
bhanupratapbiswas/bollywood-actress-name-and-movie-list & Categorical & 0.571 & 0.503 & 0.246 & 0.449 & 0.404 \\
bhanupratapbiswas/fashion-products & Categorical & 0.351 & 0.345 & 0.311 & 0.336 & 0.338 \\
bhanupratapbiswas/ipl-dataset-2008-2016 & Categorical & 0.139 & 0.939 & 0.624 & 0.593 & 0.671 \\
bhanupratapbiswas/uber-data-analysis & Categorical & 0.932 & 0.845 & 0.922 & 0.922 & 0.858 \\
bhanupratapbiswas/world-top-billionaires & Quartile & 0.267 & 0.557 & 0.245 & 0.322 & 0.456 \\
bharath011/heart-disease-classification-dataset & Binary & 0.623 & 0.612 & 0.435 & 0.668 & 0.666 \\
bhavkaur/hotel-guests-dataset & Categorical & 0.867 & 0.606 & 0.096 & 0.508 & 0.697 \\
bhavkaur/simplified-titanic-dataset & Binary & 0.749 & 0.600 & 0.523 & 0.661 & 0.555 \\
blastchar/telco-customer-churn & Binary & 0.731 & 0.691 & 0.278 & 0.722 & 0.718 \\
bretmathyer/telemedicine-used & Binary & 0.512 & 0.562 & 0.488 & 0.482 & 0.501 \\
buntyshah/auto-insurance-claims-data & Binary & 0.753 & 0.655 & 0.240 & 0.736 & 0.751 \\
burak3ergun/loan-data-set & Binary & 0.687 & 0.792 & 0.436 & 0.567 & 0.648 \\
carolzhangdc/imdb-5000-movie-dataset & Quartile & 0.288 & 0.577 & 0.233 & 0.259 & 0.518 \\
chirin/africa-economic-banking-and-systemic-crisis-data & Binary & 0.910 & 0.934 & 0.197 & 0.836 & 0.846 \\
christinestevens/cstevens-peloton-data & Categorical & 0.212 & 0.994 & 0.006 & 0.212 & 0.942 \\
cpluzshrijayan/milkquality & Categorical & 0.404 & 0.392 & 0.394 & 0.482 & 0.481 \\
crxxom/manhwa-dataset & Categorical & 0.636 & 0.886 & 0.596 & 0.618 & 0.838 \\
dansbecker/aer-credit-card-data & Binary & 0.783 & 0.725 & 0.648 & 0.658 & 0.624 \\
deependraverma13/diabetes-healthcare-comprehensive-dataset & Binary & 0.651 & 0.688 & 0.678 & 0.695 & 0.719 \\
desalegngeb/german-fintech-companies & Categorical & 0.362 & 0.832 & 0.046 & 0.527 & 0.687 \\
dileep070/heart-disease-prediction-using-logistic-regression & Binary & 0.855 & 0.782 & 0.670 & 0.726 & 0.642 \\
dsfelix/us-stores-sales & Quartile & 0.271 & 0.674 & 0.123 & 0.457 &  \\
elakiricoder/gender-classification-dataset & Binary & 0.521 & 0.732 & 0.518 & 0.844 & 0.765 \\
fedesoriano/hepatitis-c-dataset & Categorical & 0.867 & 0.855 & 0.867 & 0.868 & 0.857 \\
fedesoriano/stroke-prediction-dataset & Binary & 0.949 & 0.949 & 0.805 & 0.864 & 0.849 \\
gabrielsantello/cars-purchase-decision-dataset & Binary & 0.598 & 0.602 & 0.650 & 0.612 & 0.605 \\
geomack/spotifyclassification & Binary & 0.510 & 0.576 & 0.174 & 0.518 & 0.555 \\
gyanprakashkushwaha/laptop-price-prediction-cleaned-dataset & Quartile & 0.260 & 0.506 & 0.175 & 0.315 &  \\
hansrobertson/american-companies-profits-and-benefits-from-ai & Quartile & 0.253 & 0.246 & 0.048 & 0.280 & 0.256 \\
harishkumardatalab/medical-insurance-price-prediction & Quartile & 0.254 & 0.659 & 0.036 & 0.324 & 0.834 \\
harshitshankhdhar/imdb-dataset-of-top-1000-movies-and-tv-shows & Quartile & 0.322 & 0.406 & 0.291 & 0.338 &  \\
hashemi221022/bank-loans & Binary & 0.919 & 0.877 & 0.914 & 0.901 & 0.868 \\
hashemi221022/diabetes & Binary & 0.651 & 0.691 & 0.682 & 0.664 & 0.711 \\
hawkingcr/airbnb-for-boston-with-fraud-detection & Binary & 0.792 & 0.749 & 0.725 & 0.682 & 0.618 \\
hemanthhari/psycological-effects-of-covid & Quartile & 0.293 & 0.421 & 0.318 & 0.302 &  \\
hesh97/titanicdataset-traincsv & Binary & 0.616 & 0.763 & 0.633 & 0.676 & 0.710 \\
iamsumat/spotify-top-2000s-mega-dataset & Quartile & 0.256 & 0.449 & 0.085 & 0.288 &  \\
iqmansingh/company-employee-dataset & Categorical & 0.150 & 0.537 & 0.085 & 0.101 & 0.508 \\
ishadss/productivity-prediction-of-garment-employees & Quartile & 0.254 & 0.395 & 0.079 & 0.292 & 0.347 \\
jainilcoder/netflix-stock-price-prediction & Quartile & 0.251 & 0.982 & 0.000 & 0.388 & 0.745 \\
jillanisofttech/brain-stroke-dataset & Binary & 0.959 & 0.958 & 0.834 & 0.879 & 0.858 \\
kabure/german-credit-data-with-risk & Binary & 0.700 & 0.573 & 0.477 & 0.646 & 0.675 \\
kandij/diabetes-dataset & Binary & 0.651 & 0.698 & 0.677 & 0.680 & 0.721 \\
kanths028/usa-housing & Quartile & 0.262 & 0.276 & 0.031 & 0.258 & 0.370 \\
kingabzpro/cosmetics-datasets & Categorical & 0.217 & 0.810 & 0.213 & 0.710 & 0.768 \\
kreeshrajani/human-stress-prediction & Binary & 0.538 & 0.566 & 0.555 & 0.613 & 0.613 \\
kumargh/pimaindiansdiabetescsv & Categorical & 0.176 & 0.111 & 0.100 & 0.097 & 0.108 \\
larsen0966/student-performance-data-set & Categorical & 0.160 & 0.291 & 0.305 & 0.336 & 0.390 \\
lightonkalumba/us-womens-labor-force-participation & Binary & 0.568 & 0.996 & 0.534 & 0.703 & 0.892 \\
mahnazarjmand/bank-personal-loan & Binary & 0.916 & 0.884 & 0.904 & 0.887 & 0.872 \\
maryalebron/life-expectancy-data & Categorical & 0.038 & 0.028 & 0.033 & 0.021 & 0.023 \\
maryammanoochehry/bank-personal-loan & Binary & 0.897 & 0.870 & 0.870 & 0.864 & 0.875 \\
mathchi/diabetes-data-set & Binary & 0.651 & 0.691 & 0.697 & 0.703 & 0.720 \\
mayankpatel14/second-hand-used-cars-data-set-linear-regression & Quartile & 0.250 & 0.262 & 0.008 & 0.264 & 0.282 \\
mayurdalvi/simple-linear-regression-placement-data & Binary & 0.511 & 0.507 & 0.480 & 0.486 & 0.501 \\
mayuriawati/bangalore-chain-restaurants-ratings-and-reviews & Categorical & 0.157 & 0.885 & 0.618 & 0.818 & 0.883 \\
mazlumi/ielts-writing-scored-essays-dataset & Categorical & 0.178 & 0.218 & 0.227 & 0.199 & 0.179 \\
mfaisalqureshi/spam-email & Binary & 0.871 & 0.905 & 0.796 & 0.846 & 0.749 \\
mirichoi0218/insurance & Quartile & 0.255 & 0.602 & 0.097 & 0.339 &  \\
muhammadtsabitulazmi/liga-1-indonesia-player-dataset & Categorical & 0.160 & 0.206 & 0.002 & 0.130 & 0.136 \\
nancyalaswad90/review & Binary & 0.651 & 0.680 & 0.671 & 0.685 & 0.723 \\
naveenkumar20bps1137/predict-students-dropout-and-academic-success & Categorical & 0.167 & 0.082 & 0.109 & 0.088 & 0.082 \\
nikhil1e9/netflix-stock-price & Quartile & 0.255 & 0.990 & 0.188 & 0.360 & 0.791 \\
ninzaami/loan-predication & Binary & 0.687 & 0.785 & 0.241 & 0.622 & 0.677 \\
noordeen/insurance-premium-prediction & Quartile & 0.255 & 0.565 & 0.059 & 0.315 &  \\
oles04/bundesliga-seasons & Binary & 0.588 & 1.000 & 0.945 & 0.972 & 0.978 \\
oles04/top-leagues-player & Quartile & 0.273 & 0.678 & 0.045 & 0.290 &  \\
patelprashant/employee-attrition & Binary & 0.837 & 0.829 & 0.803 & 0.836 & 0.822 \\
pavansubhasht/ibm-hr-analytics-attrition-dataset & Binary & 0.836 & 0.832 & 0.689 & 0.833 & 0.817 \\
phangud/spamcsv & Binary & 0.864 & 0.896 & 0.726 & 0.912 & 0.761 \\
prevek18/ames-housing-dataset & Quartile & 0.271 & 0.582 & 0.304 & 0.291 &  \\
primaryobjects/voicegender & Binary & 0.501 & 0.630 & 0.269 & 0.551 & 0.598 \\
prkhrawsthi/bitcoin-usd-daily-price-with-volume-2015-2023 & Quartile & 0.283 & 0.990 & 0.098 & 0.323 & 0.665 \\
raddar/icr-integer-data & Binary & 0.825 & 0.731 & 0.762 & 0.778 & 0.363 \\
rajyellow46/wine-quality & Categorical & 0.437 & 0.389 & 0.362 & 0.368 & 0.333 \\
ravibarnawal/mutual-funds-india-detailed & Categorical & 0.299 & 0.236 & 0.240 & 0.259 & 0.291 \\
receplyasolu/6k-weather-labeled-spotify-songs & Categorical & 0.280 & 0.166 & 0.000 & 0.145 & 0.127 \\
redwankarimsony/heart-disease-data & Categorical & 0.447 & 0.359 & 0.228 & 0.321 & 0.310 \\
reihanenamdari/breast-cancer & Quartile & 0.273 & 0.296 & 0.260 & 0.271 & 0.298 \\
rishikeshkonapure/hr-analytics-prediction & Binary & 0.839 & 0.826 & 0.800 & 0.840 & 0.814 \\
rkiattisak/student-performance-in-mathematics & Quartile & 0.266 & 0.543 & 0.200 & 0.496 & 0.569 \\
rounakbanik/pokemon & Binary & 0.913 & 0.981 & 0.964 & 0.966 & 0.975 \\
rpaguirre/tesla-stock-price & Quartile & 0.261 & 0.985 & 0.076 & 0.330 & 0.816 \\
rtatman/chocolate-bar-ratings & Categorical & 0.221 & 0.163 & 0.220 & 0.200 & 0.171 \\
ruchi798/student-feedback-survey-responses & Categorical & 0.111 & 0.104 & 0.098 & 0.104 & 0.099 \\
ruchi798/tv-shows-on-netflix-prime-video-hulu-and-disney & Categorical & 0.395 & 0.354 & 0.202 & 0.223 & 0.238 \\
sabasaeed1953/stock-prices-of-2023 & Quartile & 0.250 & 0.980 & 0.070 & 0.366 & 0.623 \\
saloni1712/chatgpt-app-reviews & Categorical & 0.520 & 0.681 & 0.660 & 0.642 & 0.690 \\
sanjanchaudhari/bankloan & Binary & 0.629 & 0.610 & 0.330 & 0.612 & 0.582 \\
sanjanchaudhari/netflix-dataset & Categorical & 0.325 & 0.534 & 0.428 & 0.452 & 0.565 \\
sanjanchaudhari/user-behavior-on-instagram & Binary & 0.507 & 0.583 & 0.492 & 0.487 & 0.488 \\
saunakghosh/nba-players-dataset & Categorical & 0.750 & 0.772 & 0.556 & 0.651 & 0.601 \\
saurabh00007/diabetescsv & Binary & 0.651 & 0.714 & 0.664 & 0.698 & 0.737 \\
sbhatti/financial-sentiment-analysis & Categorical & 0.542 & 0.641 & 0.099 & 0.694 & 0.756 \\
shashankshukla123123/marketing-campaign & Binary & 0.844 & 0.771 & 0.565 & 0.819 & 0.690 \\
shivamb/disney-movies-and-tv-shows & Binary & 0.727 & 0.986 & 0.639 & 0.895 & 0.938 \\
shivamb/hm-stores-dataset & Categorical & 0.557 & 0.483 & 0.165 & 0.548 & 0.472 \\
shreyanshverma27/imdb-horror-chilling-movie-dataset & Quartile & 0.293 & 0.537 & 0.145 & 0.299 & 0.402 \\
shreyapurohit/anime-data & Quartile & 0.269 & 0.777 & 0.009 & 0.374 &  \\
shroukgomaa/babies-food-ingredients & Categorical & 0.339 & 0.293 & 0.300 & 0.286 & 0.343 \\
shubhamgupta012/titanic-dataset & Binary & 0.618 & 0.754 & 0.668 & 0.642 & 0.690 \\
siddharthss/crop-recommendation-dataset & Categorical & 0.057 & 0.212 & 0.003 & 0.057 & 0.080 \\
sidhus/crab-age-prediction & Categorical & 0.153 & 0.109 & 0.102 & 0.117 & 0.097 \\
sudarshan6561/ipl-2023 & Categorical & 0.357 & 0.637 & 0.060 & 0.521 & 0.669 \\
suraj520/dairy-goods-sales-dataset & Quartile & 0.268 & 0.631 & 0.244 & 0.286 &  \\
surajjha101/stores-area-and-sales-data & Quartile & 0.250 & 0.250 & 0.007 & 0.305 & 0.276 \\
surajjha101/top-youtube-channels-data & Categorical & 0.241 & 0.504 & 0.305 & 0.445 & 0.504 \\
swathiunnikrishnan/amazon-consumer-behaviour-dataset & Quartile & 0.331 & 0.286 & 0.221 & 0.297 & 0.324 \\
tarkkaanko/amazon & Categorical & 0.775 & 0.823 & 0.798 & 0.762 & 0.786 \\
team-ai/spam-text-message-classification & Binary & 0.863 & 0.897 & 0.739 & 0.903 & 0.761 \\
teertha/ushealthinsurancedataset & Quartile & 0.256 & 0.617 & 0.046 & 0.331 &  \\
tejashvi14/employee-future-prediction & Binary & 0.650 & 0.544 & 0.540 & 0.545 & 0.566 \\
tejashvi14/engineering-placements-prediction & Binary & 0.556 & 0.631 & 0.666 & 0.661 & 0.694 \\
thedevastator/cancer-patients-and-air-pollution-a-new-link & Categorical & 0.077 & 0.057 & 0.053 & 0.069 & 0.055 \\
thedevastator/employee-attrition-and-factors & Binary & 0.833 & 0.830 & 0.724 & 0.834 & 0.818 \\
thedevastator/higher-education-predictors-of-student-retention & Binary & 0.867 & 0.861 & 0.835 & 0.793 & 0.779 \\
therealsampat/predict-movie-success-rate & Categorical & 0.821 & 0.806 & 0.772 & 0.797 & 0.781 \\
timoboz/tesla-stock-data-from-2010-to-2020 & Quartile & 0.268 & 0.988 & 0.000 & 0.399 & 0.765 \\
uciml/indian-liver-patient-records & Binary & 0.714 & 0.616 & 0.511 & 0.583 & 0.554 \\
uciml/mushroom-classification & Binary & 0.511 & 0.767 & 0.097 & 0.557 & 0.660 \\
uciml/pima-indians-diabetes-database & Binary & 0.651 & 0.684 & 0.673 & 0.681 & 0.728 \\
uciml/red-wine-quality-cortez-et-al-2009 & Categorical & 0.420 & 0.421 & 0.291 & 0.393 & 0.405 \\
varpit94/tesla-stock-data-updated-till-28jun2021 & Quartile & 0.271 & 0.980 & 0.005 & 0.438 & 0.879 \\
vedavyasv/usa-housing & Quartile & 0.257 & 0.301 & 0.088 & 0.246 & 0.326 \\
vijayvvenkitesh/microsoft-stock-time-series-analysis & Quartile & 0.267 & 0.985 & 0.004 & 0.356 & 0.756 \\
vikramamin/customer-churn-decision-tree-and-random-forest & Binary & 0.725 & 0.692 & 0.289 & 0.720 & 0.716 \\
vikramamin/time-series-forecasting-using-prophet-in-r & Quartile & 0.253 & 0.365 & 0.000 & 0.337 & 0.445 \\
vstacknocopyright/blood-transfusion-service-center-data & Binary & 0.762 & 0.553 & 0.683 & 0.662 & 0.627 \\
warcoder/earthquake-dataset & Quartile & 0.303 & 0.861 & 0.144 & 0.456 &  \\
wearefuture01/hepatitis-c-prediction & Categorical & 0.867 & 0.863 & 0.857 & 0.865 & 0.841 \\
whenamancodes/predict-diabities & Binary & 0.651 & 0.681 & 0.673 & 0.699 & 0.719 \\
whenamancodes/students-performance-in-exams & Quartile & 0.275 & 0.573 & 0.099 & 0.501 &  \\
yasserh/titanic-dataset & Binary & 0.616 & 0.786 & 0.632 & 0.670 & 0.712 \\
yasserh/wine-quality-dataset & Categorical & 0.427 & 0.427 & 0.375 & 0.389 & 0.333 \\
ybifoundation/food-app-business & Quartile & 0.380 & 0.276 & 0.186 & 0.315 & 0.236 \\
\end{longtable}
}

%%%%%%%%%%%%%%%%%%%%%%%%%%%%%%%%%%%%%%%%%%%%%%%%%%%%%%%%%%%%%%%%%%%%%%%%%%%%%%%
% APPENDIX: Evaluation Gaps Are Endemic Across TLM Research
%%%%%%%%%%%%%%%%%%%%%%%%%%%%%%%%%%%%%%%%%%%%%%%%%%%%%%%%%%%%%%%%%%%%%%%%%%%%%%%
\clearpage
\section{Evaluation Gaps Are Endemic Across TLM Research}
\label{app:endemic_gaps}

A natural concern with our analysis is that it focuses on a single case study. 
To address this, we systematically examined the evaluation methodology of five other influential TLM papers. 
Our analysis provides further evidence that the gaps we identify in Tabula-8B are endemic patterns across the TLM literature.

We selected five papers representing the breadth of recent TLM research:
\vspace{-0.2in}
\begin{enumerate}[leftmargin=*,noitemsep]
    \item \textbf{~\citet{fang2024llmstabulardatasurvey} Survey}: A comprehensive survey that consolidates evaluation practices and recommends benchmark datasets.
    \item \textbf{TabLLM}~\cite{hegselmann23tabllm}: This work introduced few-shot classification via serialization and prompting, evaluating on 12 datasets.
    \item \textbf{UniPredict}~\cite{wang2023unipredict}: A paper training GPT-2 on 169 Kaggle datasets, claiming universal classification capabilities.
    \item \textbf{GTL}~\cite{sun2024scaling}: This work scales generative tabular learning to 972 datasets.
    \item \textbf{CARTE}~\cite{kim2024carte}: While not strictly being a general TLM, this work proposes context-aware representations using graph-attentional for tabular transfer learning, evaluating on 51 datasets. 
    % \acomment{Verify dataset count: 40 regression + 11 classification = 51-- VERIFIED}
\end{enumerate}
\vspace{-0.2in}
We evaluated each paper against the same criteria applied to our Tabula-8B analysis.

\subsection{Findings}

\subsubsection{Universal Absence of Trivial Baselines}

The most striking finding is the complete absence of majority-class baseline reporting across all five papers. 
None report the accuracy achievable by simply predicting the most frequent class, and consequently none report lift over this baseline. 
Additionally, none report chance-corrected metrics such as Cohen's $\kappa$. 
% Table~\ref{tab:endemic_baselines} summarizes these findings.
% \begin{table}[h]
% \caption{Baseline and metric reporting across five major TLM papers. No paper reports majority-class baselines, lift metrics, or chance-corrected statistics.}
% \label{tab:endemic_baselines}
% \begin{center}
% \begin{small}
% \begin{sc}
% \begin{tabular}{l|ccc}
% \toprule
% \textbf{Paper} & \textbf{Maj.-Class} & \textbf{Lift} & \textbf{$\kappa$/MCC} \\
% \midrule
% Fang et al.\ Survey & \ding{55} & \ding{55} & \ding{55} \\
% TabLLM & \ding{55} & \ding{55} & \ding{55} \\
% UniPredict & \ding{55} & \ding{55} & \ding{55} \\
% GTL Scaling & \ding{55} & \ding{55} & \ding{55} \\
% CARTE & \ding{55} & \ding{55} & \ding{55} \\
% \bottomrule
% \end{tabular}
% \end{sc}
% \end{small}
% \end{center}
% \vskip -0.2in
% \end{table}
This omission is consequential. 
UniPredict, for instance, reports only raw accuracy across 169 datasets with no chance correction. 
The paper claims ``a notable increase in absolute accuracy of 2.2\% when compared to XGBoost,'' but without majority-baseline context, it is impossible to determine whether this represents meaningful lift or noise on potentially imbalanced datasets. 
TabLLM uses AUC as its primary metric, which is more robust to class imbalance than accuracy, but never discusses class distributions in its evaluation datasets.

\subsubsection{Incomplete Task-Type Stratification}

While some papers separate classification from regression, none separate binary from multiclass classification, and none conduct statistical tests for heterogeneity across task types. 
% Table~\ref{tab:endemic_stratification} summarizes these findings.
% \begin{table}[h]
% \caption{Task-type stratification across five major TLM papers. No paper separates binary from multiclass classification or conducts statistical heterogeneity tests.}
% \label{tab:endemic_stratification}
% \begin{center}
% \begin{small}
% \begin{sc}
% \begin{tabular}{l|ccc}
% \toprule
% \textbf{Paper} & \textbf{Class./Reg.} & \textbf{Bin./Multi.} & \textbf{Stats} \\
% \midrule
% Fang et al.\ Survey & Partial & \ding{55} & \ding{55} \\
% TabLLM & Partial & \ding{55} & \ding{55} \\
% UniPredict & \ding{55} & \ding{55} & \ding{55} \\
% GTL Scaling & \checkmark & \ding{55} & \ding{55} \\
% CARTE & \checkmark & \ding{55} & \ding{55} \\
% \bottomrule
% \end{tabular}
% \end{sc}
% \end{small}
% \end{center}
% \vskip -0.2in
% \end{table}
UniPredict is the most problematic case, reporting only aggregate accuracy across all 169 datasets. 
The paper notes that ``for continuous numerical targets (e.g., regression), the categories are defined by their quantiles,'' converting regression to classification, but provides no separate analysis of these converted tasks versus natural classification problems.
In contrast, CARTE provides the best stratification among the papers we examined, with separate learning curves for 40 regression datasets and 11 classification datasets. 
However, even CARTE does not separate binary from multiclass classification or test whether the method's advantage varies systematically by task type.

\subsubsection{Contamination Analysis Is Universally Absent}

All five papers rely on publicly available datasets from UCI, Kaggle, and OpenML that are almost certainly present in LLM training corpora, yet none conduct contamination analysis. 
This is particularly concerning given recent evidence that LLMs have memorized canonical tabular benchmarks~\cite{bordt2024elephants,carlini2023quantifying}.

The~\citet{fang2024llmstabulardatasurvey} survey implicitly acknowledges the problem, noting that ``GPT models were trained on a significant amount of web data and thus, probably exposed to more HTML and XML formats when interpreting tables,'' but never connects this observation to evaluation validity.
The survey recommends a ``combo of 9 datasets for benchmark'' that includes UCI classics such as Adult (Income), Diabetes (Pima Indians), and Heart Disease. 
\citet{bordt2024elephants} demonstrated that GPT-3.5 and GPT-4 have memorized many popular tabular datasets verbatim.
For example, GPT-4 can consistently generate the entire Iris and Wine datasets from the UCI machine learning repository.
The authors found that LLMs perform better on datasets seen during training, indicating that memorization leads to overfitting.

TabLLM's evaluation suite also consists almost entirely of high-contamination-risk datasets, including Adult Income, German Credit, California Housing, and several other UCI classics. 
UniPredict evaluates on 169 Kaggle datasets including multiple versions of Pima Indians Diabetes, Titanic, and Wine Quality, with no contamination discussion.
CARTE uses YAGO3 (built from Wikidata and Wikipedia) for pretraining, then evaluates on datasets containing entities that likely appear in YAGO3, such as company employees and movies, without overlap analysis. This concern is supported by \citet{silvestri2025evaluating}. They found that contamination effects emerge for datasets containing strong semantic cues, such as meaningful column names or interpretable value categories.
When such cues are removed or randomized, performance drops to near-random levels.
This finding suggests that LLMs' apparent competence on tabular reasoning tasks may reflect memorization of publicly available datasets rather than genuine generalization.

The only TLM paper we identified that explicitly addresses contamination is Tabula-8B~\cite{gardner2024large}, which states ``Given that T4 consists of 4M tables sourced from public data sources (Common Crawl, Github) and that our evaluations are also comprised of public benchmarks, we investigate the extent and possible impact of data contamination.'' 
TabPFN~\cite{DBLP:conf/iclr/Hollmann0EH23} avoids the problem entirely through synthetic pretraining on procedurally-generated datasets.

\subsubsection{Instruction-Tuning Confounded with Tabular Learning}
Next, existing works inadequately control for instruction-following capability in TLM evaluation, making it unclear whether performance gains stem from genuine tabular reasoning or general language model capabilities. TabLLM uses T0 and GPT-3 (\texttt{text-davinci-002}), both of which are instruction-tuned. However, they do not include comparisons with base models (e.g., T5) that lack instruction-tuning or additional fine-tuning. Thus one is no able to distinguish whether performance comes from prior knowledge of specific datasets, general instruction-following capability, semantic understanding of feature names, or actual tabular reasoning ability.

UniPredict fine-tunes GPT-2 and uses GPT-3.5 for metadata reformatting but provides no ablation separating these capabilities. 
No comparison with raw GPT-2 or other instruction-tuned models without tabular exposure is included.

The GTL  paper provides the most informative (though still partial) control. Raw LLaMA-2 achieves approximately $\approx$0.50 AUROC (chance level) on classification tasks, while instruction-tuned Phi-3 models achieve 0.58--0.63 AUROC without GTL training.
This at least demonstrates that base models do not perform tabular classification well, but does not fully separate instruction-following from tabular pattern learning in the trained models.

\subsubsection{Reproducibility}
Table~\ref{tab:endemic_reproducibility} summarizes reproducibility practices across the five papers. CARTE demonstrates best practices for reproducibility among the papers examined, releasing both code (BSD-3-Clause license) and datasets on HuggingFace, with results reported across 10 random train/test splits including standard deviations.
TabLLM releases code but not raw predictions, and healthcare dataset code is withheld ``due to privacy concerns.'' Alternatively, UniPredict stands out negatively with no code, no predictions, no standard deviations across runs, no confidence intervals, and no statistical significance tests.
The paper evaluates on 169 datasets but provides no way to verify results or understand variance.
\begin{table}[h]
\caption{Reproducibility practices across five major TLM papers. No paper releases raw predictions enabling independent verification.}
\label{tab:endemic_reproducibility}
\begin{center}
\begin{small}
\begin{sc}
\begin{tabular}{l|ccc}
\toprule
\textbf{Paper} & \textbf{Code} & \textbf{Preds.} & \textbf{Data} \\
\midrule
Fang et al.\ Survey & \checkmark & \ding{55} & \checkmark  \\
TabLLM & \checkmark & \ding{55} & \checkmark \\
UniPredict & \ding{55} & \ding{55} & \checkmark \\
GTL Scaling & Partial & \ding{55} & \checkmark \\
CARTE & \checkmark & \ding{55} & \checkmark \\
\bottomrule
\end{tabular}
\end{sc}
\end{small}
\end{center}
\vskip -0.2in
\end{table}

\subsection{Summary}

Table~\ref{tab:endemic_summary} consolidates our findings across all evaluation dimensions. TLM papers exhibit similar methodological gaps. 
This evidence strongly supports the conclusion that the issues we identify in our Tabula-8B case study are representative of endemic field-wide issues rather than isolated oversights. These findings do not diminish the contributions of individual papers, many of which have advanced the field in important ways.
Rather, they underscore the need for field-wide adoption of the evaluation standards we propose in Section~\ref{sec:recommendations} (or at least initiate a broader discussion around what the standards should be).

\begin{table}[h]
\caption{Evaluation methodology scorecard across four other major TLM approaches.
\checkmark\ = adequately addressed; 
\textbf{$\sim$} = partially addressed; 
\ding{55} = not addressed. 
All four papers exhibit the same gaps identified in our Tabula-8B case study, indicating that these issues are endemic to the TLM literature rather than specific to any single work.}
\label{tab:endemic_summary}
\begin{center}
\begin{small}
\begin{sc}
\addtolength{\tabcolsep}{-0.2em}
\begin{tabular}{l|cccc}
\toprule
\textbf{Dimension} & \textbf{TabLLM} & \textbf{UniPredict} & \textbf{GTL} & \textbf{CARTE} \\
\midrule
Majority-class baseline & \ding{55} & \ding{55} & \ding{55} & \ding{55} \\
Chance-corrected metrics & \ding{55} & \ding{55} & \ding{55} & \ding{55} \\
Binary/Quartile/multiclass split & \ding{55} & \ding{55} & \ding{55} &  \textbf{$\sim$} \\
Task Heterogeneity analysis & \ding{55} & \ding{55} & \ding{55} & \ding{55} \\
Contamination analysis & \ding{55} & \ding{55} & \ding{55} & \ding{55} \\
Instruction-tuning controls & \ding{55} & \ding{55} & \textbf{$\sim$} & \ding{55} \\
\midrule
\textbf{Gaps identified} & 6/6 & 6/6 & 5/6 & 5/6 \\
\bottomrule
\end{tabular}
\end{sc}
\end{small}
\end{center}
\vskip -0.2in
\end{table}

\subsection{Extending the Framework for Closed-Source TLMs}

Several recent TLMs ship without training-corpus or weight disclosure (e.g., commercial systems exposed only through an API), which limits direct application of our recommendations. We sketch how the framework degrades for this setting; workarounds below are offered as in-principle approaches rather than validated procedures.

\textbf{Fully transferable.} R1 (majority-class baselines and chance-corrected metrics), R2 (task-type stratification), R5 (task auditing for tabular-reasoning requirements), R6 (instruction-tuned-baseline parity tests), and R7 (community scrutiny) all operate on per-dataset evaluation outputs and do not require model internals. A closed-source TLM that does not substantially outperform an open instruction-tuned baseline under R6 leaves the central claim that tabular pretraining drives performance weakly supported.

\textbf{Partially transferable.} R4 (contamination beyond row-level deduplication) can no longer search the training corpus directly, but three behavioral workarounds are available in principle. First, \emph{perturbation tests}, where memorized canonical datasets are altered to break surface cues and accuracy drops are interpreted as evidence of memorization. Second, \emph{time-based holdouts}, where evaluation datasets are constructed after the model's reported training-data cutoff. Third, \emph{verbatim-extraction (canary) probes} of the kind introduced by~\citet{bordt2024elephants}, who show that GPT-3.5 and GPT-4 reproduce canonical tabular datasets such as Iris and Wine verbatim; querying a closed-source endpoint for suspected near-verbatim records directly tests for the same contamination. Each probe upper-bounds rather than measures contamination and should be treated as suggestive.

\textbf{Not applicable.} R3 (release evaluation code) cannot be enforced on closed-source authors; the recommendation instead becomes a request for evaluator-side code release so that third parties can rerun the same protocol on whichever endpoints they choose.

The net effect is that closed-source TLMs can still be evaluated rigorously on five of seven recommendations, with R4 weakened but not eliminated and only R3 fully out of reach.

\subsection{Extending the Framework for Table-Synthesis Methods}

A separate but adjacent line of work uses language models to \emph{generate} synthetic tabular data rather than classify it. Representative examples include GReaT~\cite{borisov2023language}, which fine-tunes GPT-style models to produce realistic tabular records; TapTap~\cite{zhang2023taptap}, which couples generative table pre-training with downstream prediction; and Tabby~\cite{cromp2025tabby}, which augments the Transformer architecture with column-specific Mixture-of-Experts heads for structured-data synthesis. We exclude these methods from our main scorecard for two reasons. First, our focus is the validity of \emph{predictive} TLM evaluation, which is the regime where reported gains have driven the most adoption claims. Second, synthesis benchmarks rely on a different set of evaluation primitives, namely train-on-synthetic-test-on-real (TSTR) utility, fidelity metrics, near-duplicate detection, and privacy leakage, on which the six dimensions of Table~\ref{tab:endemic_summary} map awkwardly or not at all.

That said, two of our recommendations transfer by analogy. R1 (baselines) applies if ``majority-class'' is replaced with a trivial generator such as training-data resampling. R4 (contamination beyond row-level deduplication) applies more directly. Synthesis benchmarks frequently rely on canonical small datasets, and \citet{bordt2024elephants} show that LLMs memorize such datasets verbatim, so we would expect TSTR utility on these datasets to be inflated by leakage rather than reflect learned distributional fidelity. A full synthesis-side audit using adapted dimensions is a natural follow-up to our work but is out of scope for the present analysis.

\end{document}